\documentclass[11pt,a4paper]{article}
\usepackage[hyperref]{emnlp2020}
\usepackage{times}
\usepackage{latexsym}

\usepackage{url}
\usepackage{latexsym}
\usepackage{multirow}
\usepackage{tabularx}
\usepackage{amsmath}
\usepackage{amssymb}
\usepackage{afterpage}
\usepackage{wrapfig}
\usepackage{enumitem}
\usepackage{array}
\newcolumntype{L}{>{\raggedright\arraybackslash}m{0.97\linewidth}}
\usepackage{graphicx}       
\usepackage{cleveref}       
\usepackage{booktabs}       
\usepackage{amsfonts}       
\usepackage{nicefrac}       
\usepackage{microtype}      
\usepackage{float,pdflscape}
\usepackage{xcolor}
\usepackage{tikz}
\usepackage{caption}
\usepackage{subcaption}
\usepackage{tablefootnote}
\usepackage{color, colortbl}
\def\hyphenateAndTtWholeString #1{\xHyphenate#1$\wholeString\unskip}
\def\xHyphenate#1#2\wholeString {\if#1$%
    \else\transform{#1}%
    \takeTheRest#2\ofTheString\fi}

\def\takeTheRest#1\ofTheString\fi
{\fi \xHyphenate#1\wholeString}

\def\transform#1{\url{#1}\hskip 0pt plus 1pt}

\def\urlx #1{\href{#1}{\hyphenateAndTtWholeString{#1}}}
\usepackage{soul}
\usetikzlibrary{calc}
\usetikzlibrary{decorations.pathmorphing}
\definecolor{crandom}{HTML}{0173B2}
\definecolor{cshap}{HTML}{DE8F05}
\definecolor{clime}{HTML}{029E73}
\definecolor{cocc}{HTML}{D55E00}
\definecolor{csalm}{HTML}{CC78BC}
\definecolor{csall2}{HTML}{CA9161}
\definecolor{cinputm}{HTML}{FBAFE4}
\definecolor{cinputl2}{HTML}{949494}
\definecolor{cguidedm}{HTML}{ECE133}
\definecolor{cguidedl2}{HTML}{56B4E9}

\definecolor{myblue}{RGB}{204, 229, 255}
\definecolor{myred}{RGB}{255, 205, 205}
\definecolor{myyellow}{RGB}{253, 253, 153}
\definecolor{mygreen}{RGB}{179, 253, 179}

\aclfinalcopy 
\def\aclpaperid{958} 

\newcommand{\salmean}{$\textit{Saliency}^{\mu}$}
\newcommand{\salnorm}{$\textit{Saliency}^{\ell2}$}
\newcommand{\inputxmean}{$\textit{InputXGrad}^{\mu}$}
\newcommand{\inputxnorm}{$\textit{InputXGrad}^{\ell2}$}
\newcommand{\guidedmean}{$\textit{GuidedBP}^{\mu}$}
\newcommand{\guidednorm}{$\textit{GuidedBP}^{\ell2}$}
\newcommand{\occlusion}{\textit{Occlusion}}
\newcommand{\shapsamp}{\textit{ShapSampl}}
\newcommand{\lime}{\textit{LIME}}
\newcommand{\rand}{\textit{Random}}

\newcommand{\trans}{$\mathtt{Transformer}$}
\newcommand{\transrand}{$\mathtt{Transformer^{RI}}$}
\newcommand{\cnn}{$\mathtt{CNN}$}
\newcommand{\cnnrand}{$\mathtt{CNN^{RI}}$}
\newcommand{\lstm}{$\mathtt{LSTM}$}
\newcommand{\lstmrand}{$\mathtt{LSTM^{RI}}$}

\newcommand{\salscores}{$\omega_{x_i,c}^M$}

\newcommand{\property}[0]{diagnostic property}
\newcommand{\propertyplural}[0]{diagnostic properties}
\newcommand{\salmap}[0]{\ensuremath{\textrm{SD}}}

\definecolor{bad_res}{HTML}{800000}
\definecolor{gr}{HTML}{18109B}

\title{A Diagnostic Study of Explainability Techniques for Text Classification}

\author{Pepa Atanasova \text{    } Jakob Grue Simonsen \text{    } Christina Lioma \text{    } Isabelle Augenstein\\
  Department of Computer Science \\
  University of Copenhagen \\
  Denmark \\
  \texttt{\{pepa, simonsen, c.lioma, augenstein\}@di.ku.dk} \\}

\date{}

\begin{document}
\maketitle
\begin{abstract}
Recent developments in machine learning have introduced models that approach human performance at the cost of increased architectural complexity.
Efforts to make the rationales behind the models' predictions transparent have inspired an abundance of new explainability techniques. Provided with an already trained model, they compute saliency scores for the words of an input instance. However, there exists no definitive guide on (i) how to choose such a technique given a particular application task and model architecture, and (ii) the benefits and drawbacks of using each such technique. In this paper, we develop a comprehensive list of \propertyplural{} for evaluating existing explainability techniques. We then employ the proposed list to compare a set of diverse explainability techniques on downstream text classification tasks and neural network architectures. We also compare the saliency scores assigned by the explainability techniques with human annotations of salient input regions to find relations between a model's performance and the agreement of its rationales with human ones. Overall, we find that the gradient-based explanations perform best across tasks and model architectures, and we present further insights into the properties of the reviewed explainability techniques.
\end{abstract}
\section{Introduction}
\begin{figure}
\centering
\includegraphics[width=220pt]{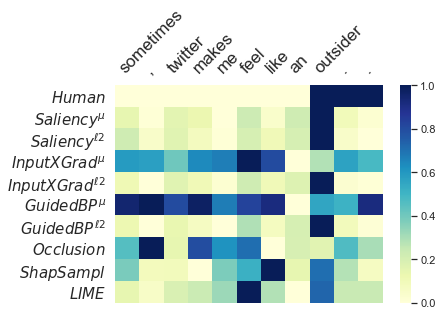}
\caption{Example of the saliency scores for the words (columns) of an instance from the Twitter Sentiment Extraction dataset. They are produced by the explainability techniques (rows) given a \trans{} model. The first row is the human annotation of the salient words. The scores are normalized in the range $[0, 1]$.}
\label{fig:example}
\end{figure}
Understanding the rationales behind models' decisions is becoming a topic of pivotal importance, as both the architectural complexity of machine learning models and the number of their application domains increases. Having greater insight into the models' reasons for making a particular prediction has already proven to be essential for discovering potential flaws or biases in medical diagnosis \cite{caruana2015intelligible} and judicial sentencing \cite{rich2016machine}. In addition, European law has mandated ``the right $\dots$ to obtain an explanation of the decision reached'' ~\cite{regulation2016regulation}.

\textit{Explainability methods} attempt to reveal the reasons behind a model's prediction for a single data point, as shown in Figure~\ref{fig:example}. They can be produced post-hoc, i.e., with already trained models. Such post-hoc explanation techniques can be applicable to one specific model~\cite{barakat2007rule, wagner2019interpretable} or to a broader range thereof~\cite{ribeiromodel, lundberg2017unified}. 
They can further be categorised as: employing model gradients \cite{sundararajan2017axiomatic, Simonyan2013DeepIC}, being perturbation based \cite{shapley1953value, zeiler2014visualizing} or providing explanations through model simplifications \cite{ribeiromodel, johansson2004truth}. There also exist explainability methods that generate textual explanations~\cite{camburu2018snli} and are trained post-hoc or jointly with the model at hand. 

While there is a growing amount of explainability methods, we find that they can produce varying, sometimes contradicting explanations, as illustrated in Figure~\ref{fig:example}.
Hence, it is important to \emph{assess existing techniques} and to \emph{provide a generally applicable and automated methodology} for choosing one that is suitable for a particular model architecture and application task~\cite{jacovi2020towards}. 
\citet{robnik2018perturbation} compiles a list of property definitions for explainability techniques, but it remains a challenge to evaluate them in practice. Several other studies have independently proposed different setups for probing varied aspects of explainability techniques~\cite{deyoung2019eraser, sundararajan2017axiomatic}.
However, existing studies evaluating explainability methods are discordant and do not compare to properties from previous studies. In our work, we consider properties from related work and extend them to be applicable to a broader range of downstream tasks.

Furthermore, to create a thorough setup for evaluating explainability methods, one should include at least: (i) different groups of explainability methods (explanation by simplification, gradient-based, etc.), (ii) different downstream tasks, and (iii) different model architectures. However, existing studies usually consider at most two of these aspects, thus providing insights tied to a specific setup.  

We propose a number of \propertyplural{} for explainability methods and evaluate them in a comparative study. We consider explainability methods from different groups, all widely applicable to most ML models and application tasks. We conduct an evaluation on three text classification tasks, which contain human annotations of salient tokens. Such annotations are available for Natural Language Processing (NLP) tasks, as they are relatively easy to obtain. This is in contrast to ML sub-fields such as image analysis, for which we only found one relevant dataset -- 536 manually annotated object bounding boxes for Visual Question Answering~\cite{subramanian2020obtaining}. 

We further compare explainability methods across three of the most widely used model architectures -- \cnn, \lstm, and \trans{}. The \trans{} model achieves state-of-the-art performance on many text classification tasks but has a complex architecture, hence methods to explain its predictions are strongly desirable. The proposed properties can also be directly applied to Machine Learning (ML) subfields other than NLP. The code for the paper is publicly available.\footnote{https://github.com/copenlu/xai-benchmark} \\In summary, the \textbf{contributions} of this work are:
\begin{itemize}[noitemsep]
\item We compile a comprehensive list of \propertyplural{} for explainability and automatic measurement of them, allowing for their effective assessment in practice.
\item We study and compare the characteristics of different groups of explainability techniques in three different application tasks and three different model architectures.
\item We study the attributions of the explainability techniques and human annotations of salient regions to compare and contrast the rationales of humans and machine learning models. 
\end{itemize}

\section{Related Work}
Explainability methods can be divided into explanations by simplification, e.g., LIME \cite{ribeiromodel}; gradient-based explanations \cite{sundararajan2017axiomatic}; perturbation-based explanations \cite{shapley1953value, zeiler2014visualizing}. Some studies propose the generation of text serving as an explanation, e.g., \cite{camburu2018snli,lei2016rationalizing,conf/acl/AtanasovaSLA20}. For extensive overviews of existing explainability approaches, see \citet{BARREDOARRIETA202082}.

Explainability methods provide explanations of different qualities, so assessing them systematically is pivotal. A common attempt to reveal shortcomings in explainability techniques is to reveal a model's reasoning process with counter-examples \cite{alvarez2018robustness, kindermans2019reliability,atanasova2020generating}, finding different explanations for the same output. However, single counter-examples do not provide a measure to evaluate explainability techniques~\cite{jacovi2020towards}.

Another group of studies performs human evaluation of the outputs of explainability methods \cite{Lertvittayakumjorn2019HumangroundedEO, narayanan2018humans}. Such studies exhibit low inter-annotator agreement and reflect mostly what appears to be reasonable and appealing to the annotators, not the actual properties of the method.

The most related studies to our work design measures and properties of explainability techniques. \citet{robnik2018perturbation} propose an extensive list of properties. The \textit{Consistency} property captures the difference between explanations of different models that produce the same prediction; and the \textit{Stability} property measures the difference between the explanations of similar instances given a single model. We note that similar predictions can still stem from different reasoning paths. Instead, we propose to explore instance activations, which reveal more of the model's reasoning process than just the final prediction. The authors propose other properties as well, which we find challenging to apply in practice. We construct a comprehensive list of \propertyplural{} tied with measures that assess the degree of each characteristic.

Another common approach to evaluate explainability methods is to measure the sufficiency of the most salient tokens for predicting the target label~\cite{deyoung2019eraser}. We also include a sufficiency estimate, but instead of fixing a threshold for the tokens to be removed, we measure the decrease of a model's performance, varying the proportion of excluded tokens. Other perturbation-based evaluation studies and measures exist~\cite{sundararajan2017axiomatic, Adebayo:2018:SCS:3327546.3327621}, but we consider the above, as it is the most widely applied.

Another direction of explainability evaluation is to compare the agreement of salient words annotated by humans to the saliency scores assigned by explanation techniques \cite{deyoung2019eraser}. We also consider the latter and further study the agreement across model architectures, downstream tasks, and explainability methods. 
While we consider human annotations at the word level \cite{camburu2018snli, lei2016rationalizing}, there are also datasets \cite{clark2019boolq,khashabi-etal-2018-looking} with annotations at the sentence-level, which would require other model architectures, so we leave this for future work.

Existing studies for evaluating explainability heavily differ in their scope. Some concentrate on a \textbf{single model architecture} - BERT-LSTM \cite{deyoung2019eraser}, RNN \cite{arras-etal-2019-evaluating}, CNN \cite{Lertvittayakumjorn2019HumangroundedEO}, whereas a few consider \textbf{more than one} model \cite{guan2019towards, poerner-etal-2018-evaluating}. Some studies concentrate on one \textbf{particular dataset} \cite{guan2019towards,arras-etal-2019-evaluating}, while only a few generalize their findings over \textbf{downstream tasks} \cite{deyoung2019eraser, vashishth2019attention}. Finally, existing studies focus on one \cite{vashishth2019attention} or a single group of explainability methods \cite{deyoung2019eraser, Adebayo:2018:SCS:3327546.3327621}. Our study is the first to propose a unified comparison of different groups of explainability techniques across three text classification tasks and three model architectures.


\section{Evaluating Attribution Maps}
We now define a set of \propertyplural{} of explainability techniques, and propose how to quantify them. Similar notions can be found in related work~\cite{robnik2018perturbation, deyoung2019eraser}, and we extend them to be generally applicable to downstream tasks. We first introduce the prerequisite notation. 
Let $X = \{(x_i, y_i, w_i)|i \in [1,N]\}$ be the test dataset, where each instance consists of a 
list of \emph{tokens} $x_i$ = $\{x_{i,j}| j \in [1, |x_i|]\}$, a \emph{gold label} $y_i$, and a 
\emph{gold saliency score} for each of the tokens in $x_i$: $w_i = \{w_{i,j} | j \in [1, |x_i|]\}$ 
with each $w_{i,j} \in \{0, 1\}$. Let $\omega$ be an explanation technique that, given a model $M$,  a class $c$, and a single instance $x_i$, computes saliency scores for each token in the input: 
\salscores $ = \{\omega_{(i,j),c}^{M} |j \in [1, |x_i|]\}$. Finally, let $M = M_1, \dots M_K$ be models with the same architecture, each trained from a randomly chosen seed, and let $M' = M_1', \dots M_K'$ be models of the same architecture, but with randomly initialized weights.

\textbf{Agreement with human rationales (HA)}. This \property{} measures the degree of overlap between saliency scores provided by human annotators, specific to the particular task, and the word saliency scores computed by an explainability technique on each instance. The property is a simple way of approximating the quality of the produced feature attributions. While it does not necessarily mean that the saliency scores explain the predictions of a model, we assume that explanations with high agreement scores would be more comprehensible for the end-user as they would adhere more to human reasoning. With this \property{}, we can also compare how the type and the performance of a model and/or dataset affect the agreement with human rationales when observing one type of explainability technique. 

During evaluation, we provide an estimate of the average agreement of the explainability technique across the dataset. To this end, we start at the instance level and compute the Average Precision (AP) of produced saliency scores \salscores{} by comparing them to the gold saliency annotations $w_i$. Here, the label for computing the saliency scores is the gold label: $c=y_i$.
Then, we compute the average across all instances, arriving at Mean AP (MAP):
\begin{equation}
\textrm{MAP}(\omega, M, X) = \frac{1}{N}\sum \limits_{i \in [1, N]} AP(w_{i}, \omega_{x_i, y_i}^M)   
\end{equation}
\textbf{Confidence Indication (CI)}. A token from a single instance can receive several saliency scores, indicating its contribution to the prediction of each of the classes. Thus, when a model recognizes a highly indicative pattern of the predicted class $k$, the tokens involved in the pattern would have highly positive saliency scores for this class and highly negative saliency scores for the remaining classes. On the other hand, when the model is not highly confident, we can assume that it is unable to recognize a strong indication of any class, and the tokens accordingly do not have high saliency scores for any class. Thus, the computed explanation of an instance $i$ should indicate the confidence $p_{i,k}$ of the model in its prediction.

We propose to measure the predictive power of the produced explanations for the confidence of the model. We start by computing the Saliency Distance (SD) between the saliency scores for the predicted class $k$ to the saliency scores of the other classes $K/k$ (Eq.~\ref{eq:conf1}). Given the distance between the saliency scores, we predict the confidence of the class with logistic regression (LR) and finally compute the Mean Absolute Error -- MAE (Eq.~\ref{eq:conf2}), of the predicted confidence to the actual one.
\begin{gather}
\textrm{SD} = \sum \limits_{j \in [0,|x|]}D(\omega_{x_{i,j}, k}^{M}, \omega_{x_{i,j}, K/k}^{M}) \label{eq:conf1}\\
\textrm{MAE}(\omega, M, X) = \sum \limits_{\substack{i \in [1, N]}} |p_{i,k} - \textrm{LR}(\salmap)|~\label{eq:conf2}
\end{gather}
For tasks with two classes, D is the subtraction of the saliency value for class k and the other class. For more than two classes, D is the concatenation of the max, min, and average across the differences of the saliency value for class k and the other classes. Low MAE indicates that model's confidence can be easily identified by looking at the produced explanations.

\textbf{Faithfulness (F)}. Since explanation techniques are employed to explain model predictions for a single instance, an essential property is that they are faithful to the model's inner workings and not based on arbitrary choices. A well-established way of measuring this property is by replacing a number of the most-salient words with a mask token~\cite{deyoung2019eraser} and observing the drop in the model's performance. To avoid choosing an unjustified percentage of words to be perturbed, we produce several dataset perturbations by masking 0, 10, 20, \dots, 100\% of the tokens in order of decreasing saliency, thus arriving at $X^{\omega^0}$, $X^{\omega^{10}}$, \dots, $X^{\omega^{100}}$. Finally, to produce a single number to measure faithfulness, we compute the area under the threshold-performance curve (AUC-TP):
\begin{equation}
\begin{aligned}
\textrm{AUC-TP}(\omega, M, X) = \\ 
\textrm{AUC}([(i, P(M(X^{\omega^0}))-M(X^{\omega^i}))]) 
\end{aligned}
\end{equation}
where P is a task specific performance measure and $i \in [0, 10, \dots, 100]$. 
We also compare the AUC-TP of the saliency methods to a random saliency map to find whether there are explanation techniques producing saliency scores without any contribution over a random score. 

Using AUC-TP, we perform an ablation analysis
which is a good approximation of whether the most salient words are also the most important ones for a model's prediction. 
However, some prior studies~\cite{feng2018pathologies} find that models remain confident about their prediction even after stripping most input tokens, leaving a few that might appear nonsensical to humans. The \propertyplural\ that follow aim to facilitate a more in-depth analysis of the alignment between the inner workings of a model and produced saliency maps.

\textbf{Rationale Consistency (RC)}.
A desirable property of an explainability technique is to be consistent with the similarities in the reasoning paths of several models on a single instance. Thus, when two reasoning paths are similar, the scores provided by an explainability technique $\omega$ should also be similar, and vice versa. Note that we are interested in similar reasoning paths as opposed to similar predictions, as the latter does not guarantee analogous model rationales. For models with diverse architectures, we expect rationales to be diverse as well and to cause low consistency. Therefore, we focus on a set of models with the same architecture, trained from different random seeds as well as the same architecture, but with randomly initialized weights. The latter would ensure that we can have model pairs $(M_s, M_p)$ with similar and distant rationales. We further claim that the similarity in the reasoning paths could be measured effectively with the distance between the activation maps (averaged across layers and neural nodes) produced by two distinct models (Eq.~\ref{eq:consist1}). The distance between the explanation scores is computed simply by subtracting the two (Eq.~\ref{eq:consist2}). Finally, we compute  Spearman's $\rho$ between the similarity of the explanation scores and the similarity of the attribution maps (Eq.~\ref{eq:consist3}).

\begin{gather}
D(M_s, M_p, x_i) = D(M_s(x_i), M_p(x_i)) \label{eq:consist1}\\
D(M_s, M_p, x_i, \omega) = D(\omega_{x_i, y_i}^{M_s}, \omega_{x_i, y_i}^{M_p}) \label{eq:consist2} \\
\begin{split}
\rho(M_s, M_p, X, \omega) = \rho(D(M_s, M_p, x_i), \\
D(M_s, M_p, x_i, \omega)| i \in [1, N] ) \label{eq:consist3}
\end{split}
\end{gather}
The higher the positive correlation is, the more consistent the attribution method would be. We choose Spearman's $\rho$ as it measures the monotonic correlation between the two variables. On the other hand, Pearson's $\rho$ measures only the linear correlation, and we can have a non-linear correlation between the activation difference and the saliency score differences.
When subtracting saliency scores and layer activations, we also take the absolute value of the vector difference as the property should be invariant to order of subtraction.
An additional benefit of the property is that low correlation scores would also help to identify explainability techniques that are not faithful to a model's rationales.

\textbf{Dataset Consistency (DC)}. 
The next \property{} is similar to the above notion of rationale consistency but focuses on consistency across instances of a dataset as opposed to consistency across different models of the same architecture. In this case, we test whether instances with similar rationales also receive similar explanations. While Rationale Consistency compares instance explanations of the same instance for different model rationales, Dataset Consistency compares explanations for pairs of instances on the same model. We again measure the similarity between instances $x_i$ and $x_j$ by comparing their activation maps, as in Eq.~\ref{eq:consistdata1}. The next step is to measure the similarity of the explanations produced by an explainability technique $\omega$, which is the difference between the saliency scores as in Eq.~\ref{eq:consistdata2}. Finally, we measure Spearman's $\rho$  between the similarity in the activations and the saliency scores as in Eq.~\ref{eq:consistdata3}. We again take the absolute value of the difference.
\begin{gather}
D(M, x_i, x_j) = D(M(x_i), M(x_j)) \label{eq:consistdata1} \\
D(M, x_i, x_j, \omega) = D(\omega_{x_i, y_i}^{M}, \omega_{x_j, y_i}^{M})
\label{eq:consistdata2} \\
\begin{split}
\rho(M, X, \omega) = \rho(D(M, x_i, x_j), \\ D(M, x_i, x_j, \omega)| i, j \in [1, N])
\end{split} 
\label{eq:consistdata3}
\end{gather}

\section{Experiments}
\subsection{Datasets}
\begin{table}
\centering
\scriptsize
\begin{tabular}{p{8mm}p{26mm}p{14mm}p{11mm}}
\toprule
\textbf{Dataset} & \textbf{Example} & \textbf{Size} & \textbf{Length} \\ \midrule
e-SNLI \newline \cite{camburu2018snli} & \textit{Premise:} An adult dressed in black \textbf{holds a stick.} \newline \textit{Hypothesis:} An adult is walking away, \textbf{empty-handed}. \newline \textit{Label}: contradiction & 549 367 Train \newline 9 842 Dev \newline 9 824 Test & 27.4 inst. \newline 5.3 expl. \\ \midrule
Movie \newline Reviews \newline \cite{zaidan-etal-2007-using} \newline & \textit{Review:} he is one of \textbf{the most exciting martial artists on the big screen}, continuing to perform his own stunts and \textbf{dazzling audiences} with his flashy kicks and punches.\newline \textit{Class:} Positive & 1 399 Train \newline 199 Dev \newline 199 Test & 834.9 inst. \newline 56.18 expl. \\ \midrule
Tweet \newline Sentiment \newline Extraction \newline (TSE)~\footnotemark & 
\textit{Tweet:} im soo \textbf{bored}...im deffo missing my music channels \newline
\textit{Class:} Negative & 
21 983 Train \newline 2 747 Dev \newline 2 748 Test & 20.5 inst. \newline 9.99 expl. \\
\bottomrule
\end{tabular}
\caption{Datasets with human-annotated saliency explanations. The \textit{Size} column presents the dataset split sizes we use in our experiments. The \textit{Length} column presents the average number of instance tokens in the test set \textit{(inst.)} and the average number of human annotated explanation tokens \textit{(expl.)}.}
\label{tab:datasets}
\end{table}
\footnotetext{\urlx{https://www.kaggle.com/c/tweet-sentiment-extraction}}

Table~\ref{tab:datasets} provides an overview of the used datasets.
For e-SNLI, models predict inference -- contradiction, neutral, or entailment -- between sentence tuples. For the Movie Reviews dataset, models predict the sentiment -- positive, negative, or neutral -- of reviews with multiple sentences. Finally, for the TSE dataset, models predict tweets' sentiment -- positive, negative, or neutral.
The e-SNLI dataset provides three dataset splits with human-annotated rationales, which we use as training, dev, and test sets, respectively. The Movie Reviews dataset provides rationale annotations for nine out of ten splits. Hence, we use the ninth split as a test and the eighth split as a dev set, while the rest are used for training. Finally, the TSE dataset only provides rationale annotations for the training dataset, and we therefore  randomly split it into 80/10/10\% chunks for training, development and testing.

\subsection{Models}  
\begin{table}[t]
\centering
\small
\begin{tabular}{lrr}
\toprule
\textbf{Model} & \textbf{Val} & \textbf{Test} \\ \midrule
\multicolumn{3}{c}{\textbf{e-SNLI}} \\
\trans & 0.897 ($\pm$0.002) & 0.892 ($\pm 0.002$) \\
\transrand & 0.167 ($\pm$0.003) & 0.167 ($\pm 0.003$)\\
\cnn & 0.773 ($\pm$0.003) & 0.768 ($\pm 0.002$)\\
\cnnrand & 0.195 ($\pm 0.038$) & 0.194 ($\pm 0.037$) \\
\lstm & 0.794 ($\pm$0.005) & 0.793 ($\pm 0.009$)\\
\lstmrand & 0.176 ($\pm 0.013$) & 0.176 ($\pm 0.000$) \\ \midrule

\multicolumn{3}{c}{\textbf{Movie Reviews}}\\
\trans & 0.859 ($\pm$0.044) & 0.856 ($\pm$0.018) \\
\transrand & 0.335 ($\pm$0.003)& 0.333 ($\pm 0.000$)\\
\cnn & 0.831 ($\pm$0.014) & 0.773 ($\pm$0.005)\\
\cnnrand & 0.343 ($\pm$0.020) & 0.333 ($\pm 0.001$) \\
\lstm & 0.614 ($\pm$0.017) & 0.567 ($\pm 0.019$)\\
\lstmrand & 0.362 ($\pm$0.030) & 0.363 ($\pm 0.041$) \\ \midrule

\multicolumn{3}{c}{\textbf{TSE}} \\
\trans & 0.772 ($\pm$0.005) & 0.781 ($\pm 0.009$) \\
\transrand &0.165 ($\pm$0.025) & 0.171 ($\pm 0.022$)\\
\cnn & 0.708 ($\pm$0.007) &  0.730 ($\pm 0.007$)\\
\cnnrand & 0.221 ($\pm$0.060) & 0.226 ($\pm 0.055$) \\
\lstm & 0.701 ($\pm$0.005) & 0.727 ($\pm 0.004$)\\
\lstmrand & 0.196 ($\pm$0.070) & 0.204 ($\pm 0.070$) \\
\bottomrule
\end{tabular}
\caption{Models' F1 score on the test and the validation datasets. The results present the average and the standard deviation of the Performance measure over five models trained from different seeds. The random versions of the models are again five models, but only randomly initialized, without training.}
\label{tab:modeleval}
\end{table}

We experiment with different commonly used base models, namely \cnn{}~\cite{fukushima1980neocognitron}, \lstm{} ~\cite{hochreiter1997long}, and the \trans{} ~\cite{vaswani2017attention} architecture BERT \cite{devlin2019bert}. The selected models allow for a comparison of the explainability techniques on diverse model architectures.  Table~\ref{tab:modeleval} presents the performance of the separate models on the datasets.

For the \cnn{} model, we use an embedding, a convolutional, a max-pooling, and a linear layer. The embedding layer is initialized with GloVe~\cite{pennington2014glove} embeddings and is followed by a dropout layer. The convolutional layer computes convolutions with several window sizes and multiple-output channels with ReLU~\cite{hahnloser2000digital} as an activation function. The result is compressed down with a max-pooling layer, passed through a dropout layer, and into a fine linear layer responsible for the prediction. The final layer has a size equal to the number of classes in the dataset.

The \lstm{} model again contains an embedding layer initialized with the GloVe embeddings. The embeddings are passed through several bidirectional LSTM layers. The final output of the recurrent layers is passed through three linear layers and a final dropout layer.

For the \trans{} model, we fine-tune the pre-trained basic, uncased language model (LM)~\cite{Wolf2019HuggingFacesTS}. The fine-tuning is performed with a linear layer on top of the LM with a size equal to the number of classes in the corresponding task. Further implementation details for all of the models, as well as their F1 scores, are presented in~\ref{appendix:A}.

\subsection{Explainability Techniques}
We select the explainability techniques to be representative of different groups -- gradient~\cite{sundararajan2017axiomatic, Simonyan2013DeepIC}, perturbation ~\cite{shapley1953value, zeiler2014visualizing} and simplification based~\cite{ribeiromodel, johansson2004truth}. 

Starting with the \textbf{gradient-based} approaches, we select \textit{Saliency}~\cite{Simonyan2013DeepIC} as many other gradient-based explainability methods build on it. It computes the gradient of the output w.r.t. the input. We also select two widely used improvements of the \textit{Saliency} technique, namely \textit{InputXGradient}~\cite{Kindermans2016InvestigatingTI}, and \textit{Guided Backpropagation}~\cite{springenberg2014striving}. InputXGradient additionally multiplies the gradient with the input and \textit{Guided Backpropagation} overwrites the gradients of ReLU functions so that only non-negative gradients are backpropagated.

From the \textbf{perturbation-based} approaches, we employ \textit{Occlusion}~\cite{zeiler2014visualizing}, which replaces each token with a baseline token (as per standard, we use the value zero) and measures the change in the output. Another popular perturbation-based technique is the \textit{Shapley Value Sampling}~\cite{castro2009polynomial}. It is based on the Shapley Values approach that computes the average marginal contribution of each word across all possible word perturbations. The Sampling variant allows for a faster approximation of Shapley Values by considering only a fixed number of random perturbations as opposed to all possible perturbations.

Finally, we select the \textbf{simplification-based} explanation technique LIME~\cite{ribeiromodel}. For each instance in the dataset, LIME trains a linear model to approximate the local decision boundary for that instance.

\textbf{Generating explanations.} 
The saliency scores from each of the explainability methods are generated for each of the classes in the dataset. As all of the gradient approaches provide saliency scores for the embedding layer (the last layer that we can compute the gradient for), we have to aggregate them to arrive at one saliency score per input token. As we found different aggregation approaches in related studies~\cite{bansal2016ask, deyoung2019eraser}, we employ the two most common methods -- 
L2 norm and averaging (denoted as $\mu$ and $\ell2$ in the explainability method names). 
\section{Results and Discussion}
We report the measures of each \property{} as well as FLOPs as a measure of the computing time used by the particular method. For all \propertyplural, we also include the randomly assigned saliency as a baseline.

\subsection{Results}
\begin{table}[t]
\centering
\fontsize{9}{9}\selectfont
\begin{tabular}{lrrr}
\toprule
\textbf{Saliency} & \textbf{e-SNLI} & \textbf{IMDB} & \textbf{TSE} \\
\midrule
\multicolumn{4}{c}{\trans}\\
\rand & 0.201 & 0.517 & 0.185  \\
\shapsamp & 0.479 & 0.481 & 0.667  \\
\lime & \textbf{0.809} & 0.604 & 0.553  \\
\occlusion & 0.523 & 0.323 & 0.556  \\
\salmean & 0.772 & 0.671 & \underline{0.707}  \\
\salnorm & 0.781 & \textbf{0.687} & 0.696   \\
\inputxmean & 0.364 & 0.432 & 0.307  \\
\inputxnorm & \underline{0.796} & \underline{0.676} & \textbf{0.754}  \\
\guidedmean & 0.468 & 0.236 & 0.287  \\
\guidednorm & 0.782 & \underline{0.676} & 0.685  \\
\midrule
\multicolumn{4}{c}{\cnn}\\
\rand & 0.209 & 0.468 & 0.384  \\
\shapsamp & 0.460 & 0.648 & 0.630  \\
\lime & 0.571 & 0.572 & \textbf{0.681} \\
\occlusion & 0.554 & 0.411 & 0.594  \\
\salmean & 0.853 & 0.712 & 0.595  \\
\salnorm & \underline{0.875} & \textbf{0.796} & 0.631  \\
\inputxmean & 0.576 & 0.662 & 0.613  \\
\inputxnorm & \textbf{0.881} & 0.759 & \underline{0.636}  \\
\guidedmean & 0.403 & 0.346 & 0.438  \\
\guidednorm & \underline{0.875} & \underline{0.788} & 0.628  \\
\midrule
\multicolumn{4}{c}{\lstm}\\
\rand & 0.166 & 0.343 & 0.225  \\
\shapsamp & 0.606 & 0.605 & 0.526  \\
\lime & 0.759 & 0.233 & 0.630 \\
\occlusion & 0.609 & 0.589 & 0.681  \\
\salmean & 0.795 & 0.568 & 0.702  \\
\salnorm & 0.800 & 0.583 & \textbf{0.704}  \\
\inputxmean & 0.432 & 0.481 & 0.441  \\
\inputxnorm & \textbf{0.820} & \textbf{0.685} & 0.693  \\
\guidedmean & 0.492 & 0.553 & 0.410  \\
\guidednorm & \underline{0.805} & \underline{0.660} & \textbf{0.720} \\
\bottomrule
\end{tabular}
\caption{Mean of the \property{} measures for all tasks and models. The best result for the particular model architecture and downstream task is in bold and the second-best is underlined.}
\label{tab:meanprop}
\end{table}
Of the three model architectures, unsurprisingly, the \trans\ model performs best, while the \cnn\ and the \lstm\ models are close in performance. It is only for the IMDB dataset that the \lstm\ model performs considerably worse than the \cnn, which we attribute to the fact that the instances contain a large number of tokens, as shown in Table~\ref{tab:datasets}. As this is not the core focus of this paper, detailed results can be found in the supplementary material.

\textbf{Overall results.} Table~\ref{tab:meanprop} presents the mean of all properties across tasks and models with all property measures normalized to be in the range [0,1]. 
We see that gradient-based explainability techniques always have the best or the second-best performance for the \propertyplural{} across all three model architectures and all three downstream tasks. Note that, \inputxmean{} and \guidedmean{}, which are computed with a mean aggregation of the scores, have some of the worst results. We conjecture that this is due to the large number of values that are averaged -- the mean smooths out any differences in the values. In contrast, the L2 norm aggregation amplifies the presence of large and small values in the vector. From the non-gradient based explainability methods, \lime{} has the best performance, where in two out of nine cases it has the best performance. It is followed by \shapsamp{} and \occlusion{}. We can conclude that the occlusion based methods overall have the worst performance according to the \propertyplural{}.

Furthermore, we see that the explainability methods achieve better performance for the e-SNLI and the TSE datasets with the \trans{} and \lstm{} architectures, whereas the results for the IMDB dataset are the worst. We hypothesize that this is due to the longer text of the input instances in the IMDB dataset. The scores also indicate that the explainability techniques have the highest \property{} measures for the \cnn{} model with the e-SNLI and the IMDB datasets, followed by the \lstm{}, and the \trans{} model. We suggest that the performance of the explainability tools can be worse for large complex architectures with a huge number of neural nodes, like the \trans{} one, and perform better for small, linear architectures like the \cnn{}.
\begin{figure}
\centering
\begin{subfigure}{.5\textwidth}
  \centering
  \includegraphics[width=215pt]{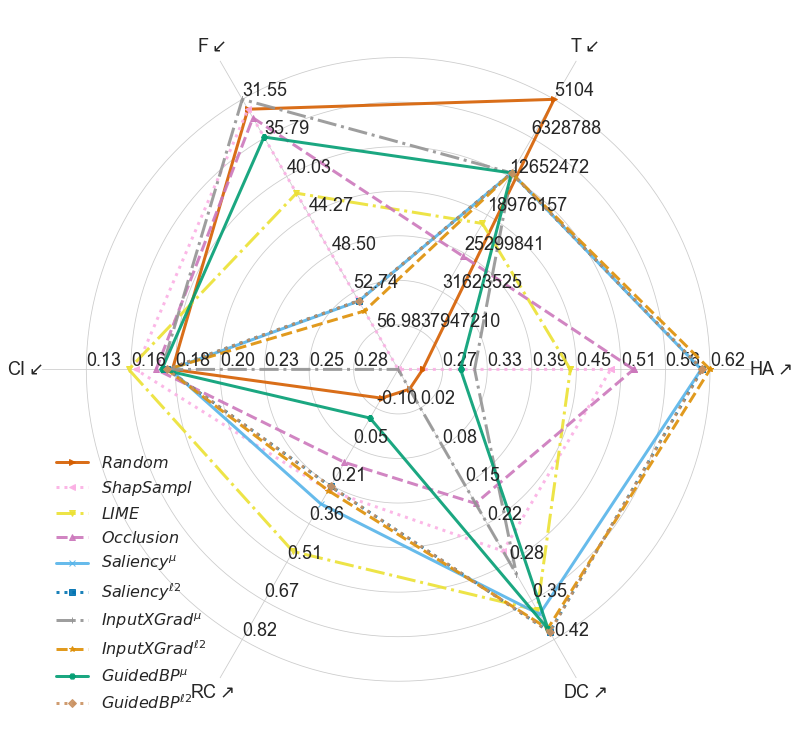}
  \caption{\trans}
  \label{fig:sub11}
\end{subfigure} 
\begin{subfigure}{.5\textwidth}
  \centering
  \includegraphics[width=215pt]{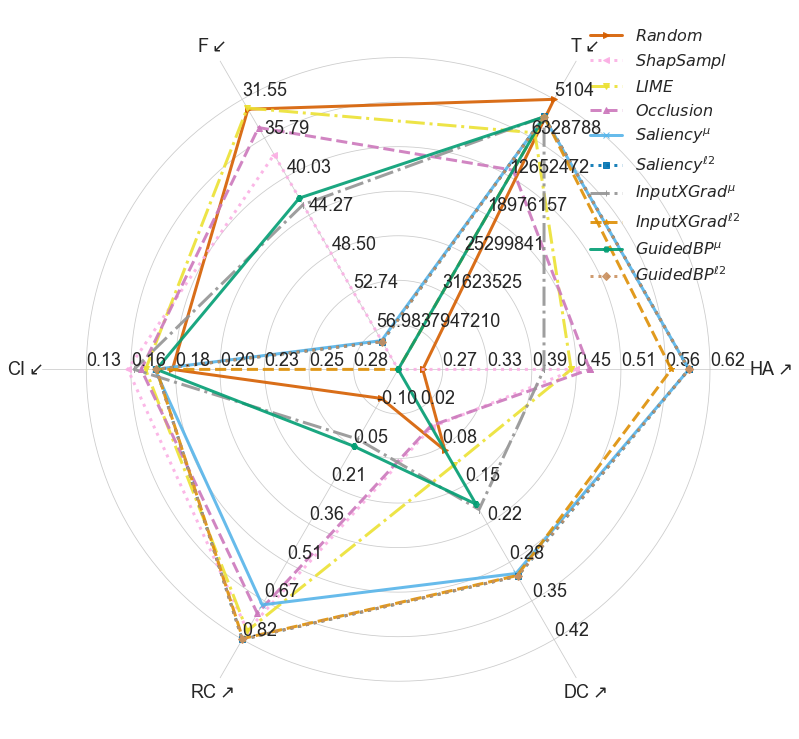}
  \caption{\cnn}
  \label{fig:sub12}
\end{subfigure}
\begin{subfigure}{.5\textwidth}
  \centering
  \includegraphics[width=215pt]{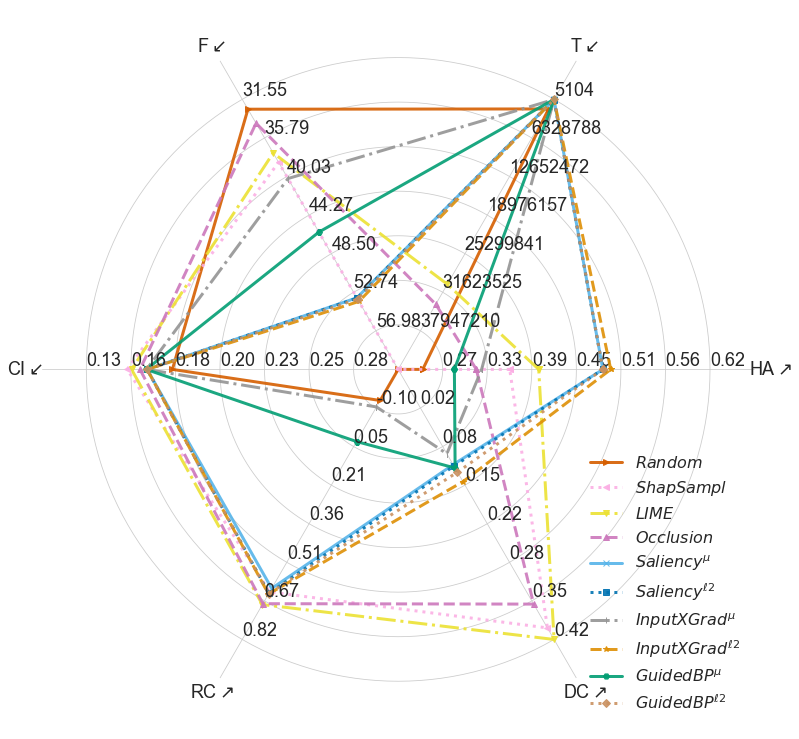}
  \caption{\lstm}
  \label{fig:sub13}
\end{subfigure}
\caption{Diagnostic property evaluation for 
all explainability techniques, on the e-SNLI dataset. 
The $\nearrow$ and $\swarrow$ signs indicate that higher, correpspondingly lower, values of the property measure are better.}
\label{fig:spider1}
\end{figure}

\begin{figure}
\centering
\begin{subfigure}{.5\textwidth}
  \centering
  \includegraphics[width=215pt]{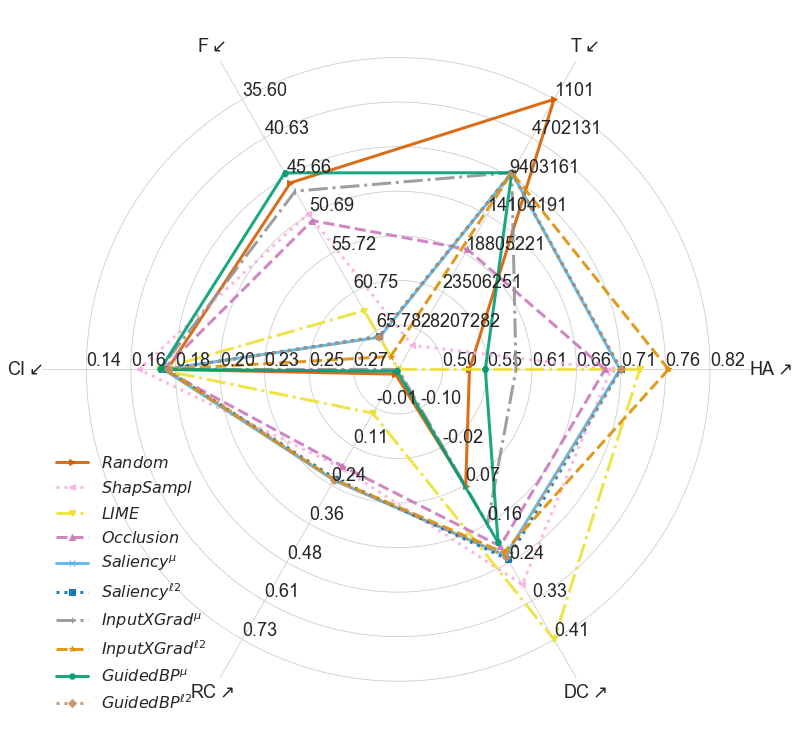}
  \caption{\trans}
  \label{fig:sub31}
\end{subfigure} 
\begin{subfigure}{.5\textwidth}
  \centering
  \includegraphics[width=215pt]{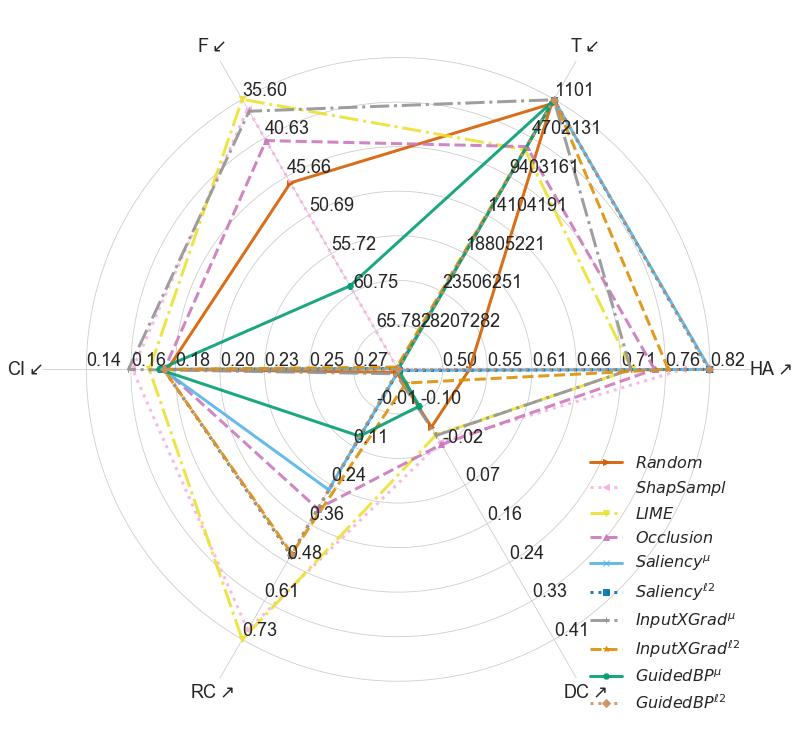}
  \caption{\cnn}
  \label{fig:sub32}
\end{subfigure}
\begin{subfigure}{.5\textwidth}
  \centering
  \includegraphics[width=215pt]{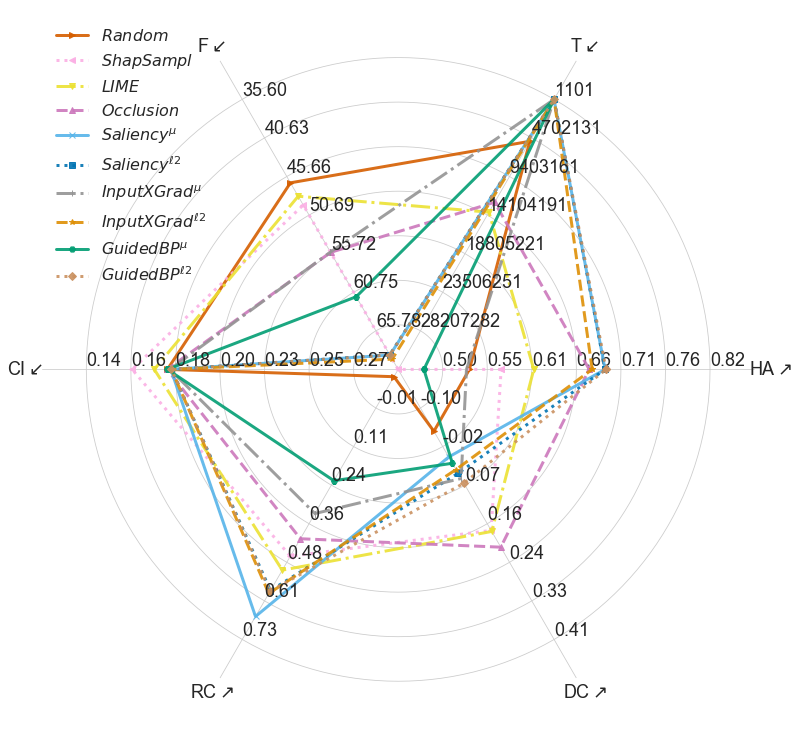}
  \caption{\lstm}
  \label{fig:sub33}
\end{subfigure}
 \caption{Diagnostic property evaluation for 
all explainability techniques, on the TSE dataset. 
The $\nearrow$ and $\swarrow$ signs indicate that higher, correspondingly lower, values of the property measure are better.}
\label{fig:spider3}
\end{figure}

\textbf{Diagnostic property performance.} Figure~\ref{fig:spider1} shows the performance of each explainability technique for all \propertyplural\ on the e-SNLI dataset, and Figure~\ref{fig:spider3} -- for the TSE dataset, which are considerably bigger than IMDB.  The IMDB dataset shows similar tendencies and a corresponding figure can be found in the supplementary material.

\textbf{Agreement with human rationales.}
We observe that the best performing explainability technique for the \trans{} model is \inputxnorm{} followed by the gradient-based ones with L2 norm aggregation.
While for the \cnn{} and the \lstm{} models, we observe similar trends, their MAP scores are always lower than for the \trans, which indicates a correlation between the performance of a model and its agreement with human rationales. 
Furthermore, the MAP scores of the \cnn{} model are higher than for the \lstm{} model, even though the latter achieves higher F1 scores on the e-SNLI dataset.
This might indicate that the representations of the \lstm{} model are less in line with human rationales.
Finally, we note that the mean aggregations of the gradient-based explainability techniques have MAP scores close to or even worse than those from the randomly initialized models.

\textbf{Faithfulness.} 
We find that gradient-based techniques have the best performance for the Faithfulness \property. On the e-SNLI dataset, it is particularly \inputxnorm{}, which performs well across all model architectures. We further find that the \cnn{} exhibits the highest Faithfulness scores for seven out of nine explainability methods. We hypothesize that this is due to the simple architecture with relatively few neural nodes compared to the recurrent nature of the \lstm{} model and the large number of neural nodes in the \trans{} architecture. Finally, models with high Faithfulness scores do not necessarily have high Human agreement scores and vice versa. This suggests that these two are indeed separate \propertyplural{}, and the first should not be confused with estimating the faithfulness of the techniques.

\textbf{Confidence Indication.} 
We find that the Confidence Indication of all models is predicted most accurately by the \shapsamp{}, \lime{}, and \occlusion{} explainability methods. This result is expected, as they compute the saliency of words based on differences in the model's confidence using different instance perturbations. We further find that the \cnn{} model's confidence 
is better predicted with \inputxmean{}. The lowest MAE with the balanced dataset is for the \cnn{} and \lstm{} models. We hypothesize that this could be due to these models' overconfidence, which makes it challenging to detect when the model is not confident of its prediction.


\textbf{Rationale Consistency.} 
There is no single universal explainability technique that achieves the highest score for Rationale Consistency property. 
We see that \lime{} can be good at achieving a high performance, which is expected, as it is trained to approximate the model's performance. The latter is beneficial, especially for models with complex architectures like the \trans. The gradient-based approaches also have high Rationale Consistency scores. We find that the \occlusion{} technique is the best performing for the \lstm{} across all tasks, as it is the simplest of the explored explainability techniques, and does not inspect the model's internals or try to approximate them. This might serve as an indication that \lstm{} models, due to their recurrent nature, can be best explained with simple perturbation based methods that do not examine a model's reasoning process. 

\textbf{Dataset Consistency.} Finally, the results for the Dataset Consistency property show low to moderate correlations of the explainability techniques with similarities across instances in the dataset. The correlation is present for LIME and the gradient-based techniques, again with higher scores for the L2 aggregated gradient-based methods. 

\textbf{Overall.} To summarise, the proposed list of \propertyplural{} allows for assessing existing explainability techniques from different perspectives and supports the choice of the best performing one. Individual property results indicate that gradient-based methods have the best performance.
The only strong exception to the above is the better performance of \shapsamp{} and \lime{} for the Confidence Indication \property. However, \shapsamp{}, \lime{} and \occlusion{} take considerably more time to compute and have worse performance for all other \propertyplural. 

\section{Conclusion}
We proposed a comprehensive list of \propertyplural{} for the evaluation of explainability techniques from different perspectives. We further used them to compare and contrast different groups of explainability techniques on three downstream tasks and three diverse architectures. We found that gradient-based explanations are the best for all of the three models and all of the three downstream text classification tasks that we consider in this work. Other explainability techniques, such as \shapsamp{}, \lime{} and \occlusion{} take more time to compute, and are in addition considerably less faithful to the models and less consistent with the rationales of the models and similarities in the datasets. 

\section*{Acknowledgements}
$\begin{array}{l}\includegraphics[width=1cm]{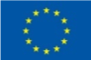} \end{array}$ This project has received funding from the European Union's Horizon 2020 research and innovation programme under the Marie Sk\l{}odowska-Curie grant agreement No 801199.





\bibliography{acl2019}
\bibliographystyle{acl_natbib}

\clearpage
\appendix
\section{Appendices}
\subsection{Experimental Setup}~\label{appendix:A}
\begin{table}[h!]
\centering
\small
\begin{tabular}{lrr}
\toprule
\textbf{Model} & \textbf{Time} & \textbf{Score} \\ \midrule
\multicolumn{3}{c}{\textbf{e-SNLI}} \\
\trans & 244.763 ($\pm$62.022) & 0.523 ($\pm$0.356) \\
\cnn & 195.041 ($\pm$53.994) & 0.756 ($\pm$0.028) \\
\lstm & 377.180 ($\pm$232.918) & 0.708 ($\pm$0.205)\\

\multicolumn{3}{c}{\textbf{Movie Reviews}}\\
\trans & 3.603 ($\pm$0.031) & 0.785 ($\pm$0.226) \\
\cnn & 4.777 ($\pm$1.953) & 0.756 ($\pm$0.058)\\
\lstm & 5.344 ($\pm$1.593) & 0.584 ($\pm$0.061) \\

\multicolumn{3}{c}{\textbf{TSE}} \\
\trans & 9.393 ($\pm$1.841) & 0.783 ($\pm$0.006) \\
\cnn & 2.240 ($\pm$0.544) & 0.730 ($\pm$0.035) \\
\lstm & 3.781 ($\pm$1.196) & 0.713 ($\pm$0.076) \\
\bottomrule
\end{tabular}
\caption{Hyper-parameter tuning details. \textit{Time} is the average time (mean and standard deviation in brackets) measured in minutes required for a particular model with all hyper-parameter combinations. \textit{Score} is the mean and standard deviation of the performance on the validation set as a function of the number of the different hyper-parameter searches.}
\label{tab:modeleval}
\end{table}

\paragraph{Machine Learning Models}. The models used in our experiments are trained on the training splits, and the parameters are selected according to the development split. We conducted fine-tuning in a grid-search manner with the ranges and parameters we describe next. We use superscripts to indicate when a parameter value was selected for one of the datasets e-SNLI -- 1, Movie Review -- 2, and TSE -- 3. For the \cnn{} model, we experimented with the following parameters: embedding dimension $\in \{50, 100, 200, 300^{1, 2, 3}\}$, batch size $\in \{16^{2}, 32, 64^{3}, 128, 256^{1}\}$, dropout rate $\in \{0.05^{1,2,3}, 0.1, 0.15, 0.2\}$, learning rate for an Adam optimizer $\in \{0.01, 0.03, 0.001^{2, 3}, 0.003, 0.0001^{1}, 0.0003\}$, window sizes $\in \{[2, 3, 4]^{2}, [2, 3, 4, 5], [3, 4, 5]^{3}, [3, 4, 5, 6],$ $[4, 5, 6], [4, 5, 6, 7]^{1}\}$, and number of output channels $\in \{50^{2, 3}, 100, 200, 300^{1}\}$. We leave the stride and the padding parameters to their default values -- one and zero. 

For the \lstm{} model we fine-tuned over the following grid of parameters: embedding dimension $\in \{50^{3}, 100^{1, 2}, 200, 300\}$, batch size $\in \{16^{3}, 32, 64, 128^{2}, 256^{1}\}$, dropout rate $\in \{0.05^{3}, 0.1^{1, 2}, 0.15, 0.2\}$, learning rate for an Adam optimizer $\in \{0.01^{1}, 0.03^{2}, 0.001^{2, 3}, 0.003, 0.0001, 0.0003\}$, number of LSTM layers $\in \{1^{2, 3}, 2, 3, 4^{1}\}$, LSTM hidden layer size $\in \{50. 100^{1, 2, 3}, 200, 300\}$, and size of the two linear layers $\in \{[50, 25]^{2}, [100, 50]^{1}, [200, 100]^{3}\}$. We also experimented with other numbers of linear layers after the recurrent ones, but having three of them, where the final was the prediction layer, yielded the best results. 

The \cnn{} and \lstm{} models are trained with an early stopping over the validation accuracy with a patience of five and a maximum number of training epochs of 100. We also experimented with other optimizers, but none yielded improvements.

Finally, for the \trans{} model we fine-tuned the pre-trained basic, uncased LM~\cite{Wolf2019HuggingFacesTS}(110M parameters) where the maximum input size is 512, and the hidden size of each layer of the 12 layers is 768. We performed a grid-search over learning rate of $\in \{1e-5, 2e-5^{1, 2}, 3e-5^{3}, 4e-5, 5e-5\}$. The models were trained with a warm-up period where the learning rate increases linearly between 0 and 1 for 0.05\% of the steps found with a grid-search. We train the models for five epochs with an early stopping with patience of one as the Transformer models are easily fine-tuned for a small number of epochs.

All experiments were run on a single NVIDIA TitanX GPU with 8GB, and 4GB of RAM and 4 Intel Xeon Silver 4110 CPUs.

The models were evaluated with macro F1 score, which can be found here \url{https://scikit-learn.org/stable/modules/generated/sklearn.metrics.precision_recall_fscore_support.html} and is defined as follows:
\begin{equation*}
   Precision (P) = \frac{\mathrm{TP}}{\mathrm{TP} + \mathrm{FP}} 
\end{equation*}
\begin{equation*}
   Recall (R) = \frac{\mathrm{TP}}{\mathrm{TP} + \mathrm{FN}} 
\end{equation*}
\begin{equation*}
   F1 = \frac{2*\mathrm{P}*\mathrm{R}}{\mathrm{P}+\mathrm{R}} 
\end{equation*}
where TP is the number of true positives, FP is the number of false positives, and FN is the number of false negatives.

\textbf{Explainability generation}. When evaluating the Confidence Indication property of the explainability measures, we train a logistic regression for 5 splits and provide the MAE over the five test splits. As for some of the models, e.g. \trans{}, the confidence is always very high, the LR starts to predict only the average confidence. To avoid this, we additionally randomly up-sample the training instances with a smaller confidence, making the number of instances in each confidence interval [0.0-0.1],\dots[0.9-1.0]) to be the same as the maximum number of instances found in one of the separate intervals.

For both Rationale and Dataset Consistency properties, we consider Spearman's $\rho$. While 
Pearson's $\rho$ measures only the linear correlation between two variables (a change in one variable should be proportional to the change in the other variable), Spearman's $\rho$ measures the monotonic correlation (when one variable increases, the other increases, too). In our experiments, we are interested in the monotonic correlation as all activation differences don't have to be linearly proportional to the differences of the explanations and therefore measure Spearman's $\rho$. 

The Dataset Consistency property is estimated over instance pairs from the test dataset. As computing it for all possible pairs in the dataset is computationally expensive, we select 2 000 pairs from each dataset in order of their decreasing word overlap and sample 2 000 from the remaining instance pairs. This ensures that we compute the \property\ on a set containing tuples of similar and different instances. 

Both the Dataset Consistency property and the Rationale Consistency property estimate the difference between the instances based on their activations. For the \lstm\ model, the activations of the LSTM layers are limited to the output activation also used for prediction as it isn't possible to compare activations with different lengths due to the different token lengths of the different instances. We also use min-max scaling of the differences in the activations and the saliencies as the saliency scores assigned by some explainability techniques are very small. 


\subsection{Spider Figure for the IMDB dataset}
~\label{appendix:C}
\begin{figure}
\centering
\begin{subfigure}{.5\textwidth}
  \centering
  \includegraphics[width=215pt]{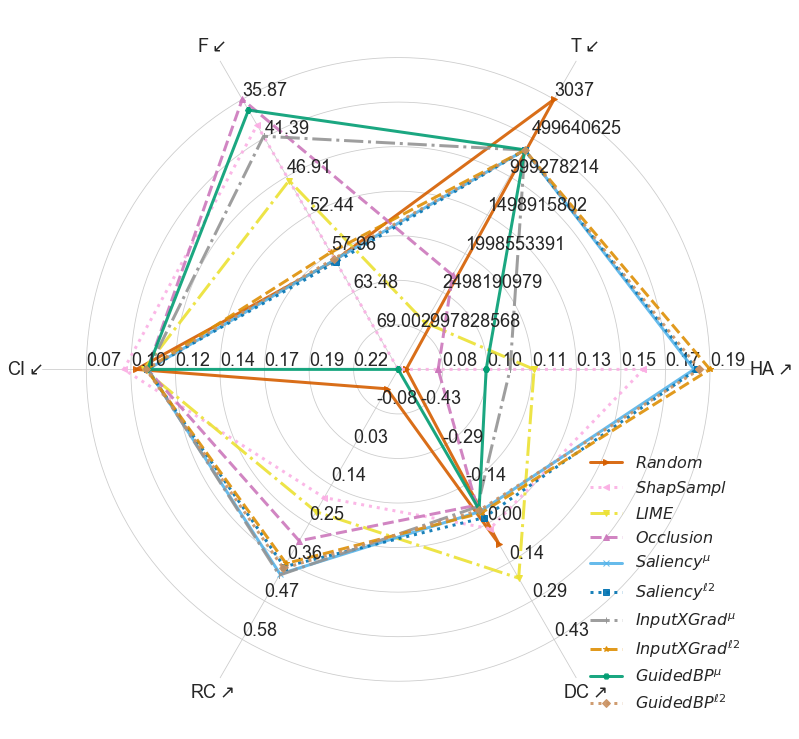}
  \caption{\trans}
  \label{fig:sub21}
  
\end{subfigure} 
\begin{subfigure}{.5\textwidth}
  \centering
  \includegraphics[width=215pt]{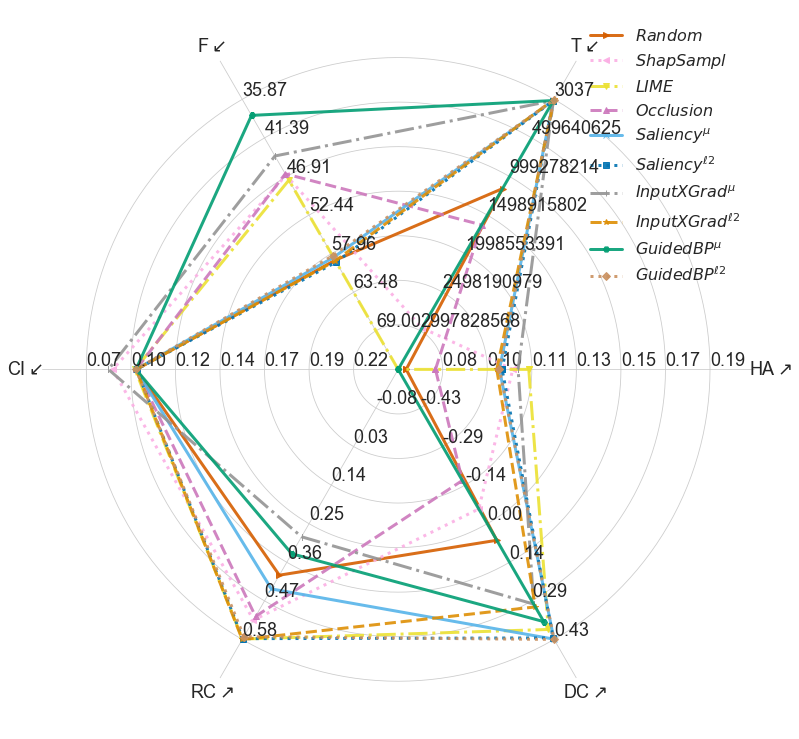}
  \caption{\cnn}
  \label{fig:sub22}
\end{subfigure}
\begin{subfigure}{.5\textwidth}
  \centering
  \includegraphics[width=215pt]{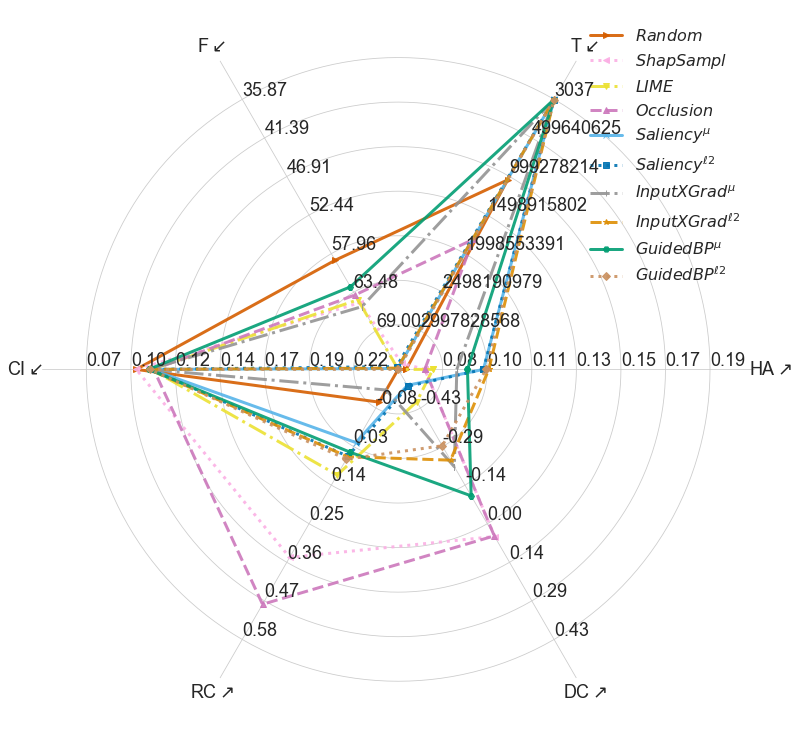}
  \caption{\lstm}
  \label{fig:sub23}
\end{subfigure}
\caption{Diagnostic property evaluation for 
all explainability techniques, on the IMDB dataset. 
The $\nearrow$ and $\swarrow$ signs following the names of each explainability method indicate that higher, correspondingly lower, values of the property measure are better.}
\label{fig:spider2}
\end{figure}

\subsection{Detailed explainability techniques evaluation results.}
~\label{appendix:B}
\begin{landscape}
\begin{table}
\centering
\footnotesize
\begin{tabular}{l@{\hspace{0.7\tabcolsep}}|r@{\hspace{0.7\tabcolsep}}r@{\hspace{0.7\tabcolsep}}l@{\hspace{0.5\tabcolsep}}|r@{\hspace{0.7\tabcolsep}}r@{\hspace{0.7\tabcolsep}}l@{\hspace{0.5\tabcolsep}}|r@{\hspace{0.7\tabcolsep}}r@{\hspace{0.7\tabcolsep}}l@{\hspace{0.5\tabcolsep}}}
\toprule
\textbf{Explain.}&\multicolumn{3}{c}{\textbf{e-SNLI}}&\multicolumn{3}{c}{\textbf{IMDB}}&\multicolumn{3}{c}{\textbf{TSE}}\\
&\textbf{MAP}&\textbf{MAP RI}&\textbf{FLOPs}&\textbf{MAP}&\textbf{MAP RI}&\textbf{FLOPs}&\textbf{MAP}&\textbf{MAP RI}&\textbf{FLOPs}\\
\midrule
\rand&.297 ($\pm$.001)&--&6.12e+3 ($\pm$4.6e+1)&.079 ($\pm$.001)&--&9.41e+4 ($\pm$1.8e+2)&.573 ($\pm$.001)&--&4.62e+3 ($\pm$2.2e+1)\\
\midrule
\multicolumn{10}{c}{\trans}\\
\shapsamp&.511 ($\pm$.004)&.292 ($\pm$.011)&1.78e+7 ($\pm$5.5e+5)&.168 ($\pm$.003)&.084 ($\pm$.001)&3.00e+9 ($\pm$1.3e+8)&.716 ($\pm$.003)&.575 ($\pm$.027)&1.29e+7 ($\pm$2.0e+6)\\
\lime&.465 ($\pm$.008)&.264 ($\pm$.004)&2.39e+5 ($\pm$1.5e+4)&.127 ($\pm$.004)&.075 ($\pm$.004)&4.98e+8 ($\pm$1.4e+8)&.745 ($\pm$.003)&.570 ($\pm$.028)&2.82e+7 ($\pm$1.6e+6)\\
\occlusion&.537 ($\pm$.014)&.292 ($\pm$.009)&6.33e+5 ($\pm$1.0e+3)&.091 ($\pm$.001)&.084 ($\pm$.001)&8.05e+7 ($\pm$4.5e+5)&.710 ($\pm$.008)&.577 ($\pm$.012)&5.86e+5 ($\pm$1.6e+2)\\
\salmean&.614 ($\pm$.003)&.255 ($\pm$.008)&5.38e+4 ($\pm$1.8e+2)&.187 ($\pm$.005)&.079 ($\pm$.001)&6.59e+5 ($\pm$1.8e+3)&.725 ($\pm$.011)&.499 ($\pm$.002)&4.93e+4 ($\pm$2.1e+2)\\
\salnorm&.615 ($\pm$.003)&.255 ($\pm$.009)&5.39e+4 ($\pm$1.3e+2)&.188 ($\pm$.006)&.078 ($\pm$.001)&6.62e+5 ($\pm$8.4e+2)&.726 ($\pm$.014)&.498 ($\pm$.001)&4.93e+4 ($\pm$1.4e+2)\\
\inputxmean&.356 ($\pm$.005)&.280 ($\pm$.016)&5.38e+4 ($\pm$1.8e+2)&.118 ($\pm$.003)&.083 ($\pm$.001)&6.60e+5 ($\pm$4.5e+3)&.620 ($\pm$.008)&.558 ($\pm$.011)&4.92e+4 ($\pm$1.4e+2)\\
\inputxnorm&\underline{\textbf{.624 ($\boldsymbol \pm$.004)}}&.254 ($\pm$.013)&5.39e+4 ($\pm$1.5e+2)&\underline{\textbf{.193 ($\boldsymbol \pm$.005)}}&.079 ($\pm$.001)&6.62e+5 ($\pm$2.1e+3)&\textbf{.774 ($\boldsymbol \pm$.009)}&.499 ($\pm$.005)&4.92e+4 ($\pm$8.0e+1)\\
\guidedmean&.340 ($\pm$.012)&.281 ($\pm$.025)&5.39e+4 ($\pm$1.8e+2)&.109 ($\pm$.003)&.086 ($\pm$.005)&6.54e+5 ($\pm$7.5e+3)&.589 ($\pm$.006)&.567 ($\pm$.008)&4.94e+4 ($\pm$4.1e+2)\\
\guidednorm&.615 ($\pm$.003)&.255 ($\pm$.009)&5.38e+4 ($\pm$1.1e+2)&.189 ($\pm$.005)&.079 ($\pm$.001)&6.59e+5 ($\pm$2.8e+3)&.726 ($\pm$.012)&.498 ($\pm$.001)&4.97e+4 ($\pm$4.2e+2)\\
\midrule
\multicolumn{10}{c}{\cnn}\\
\shapsamp&.471 ($\pm$.003)&.298 ($\pm$.008)&3.79e+7 ($\pm$3.1e+3)&.119 ($\pm$.004)&.084 ($\pm$.001)&1.26e+7 ($\pm$1.6e+5)&.789 ($\pm$.004)&.586 ($\pm$.017)&4.53e+6 ($\pm$2.1e+4)\\
\lime&.466 ($\pm$.002)&.300 ($\pm$.017)&1.81e+4 ($\pm$1.2e+3)&\textbf{.125 ($\boldsymbol \pm$.005)}&.079 ($\pm$.004)&5.39e+7 ($\pm$1.9e+4)&.737 ($\pm$.002)&.581 ($\pm$.021)&1.52e+4 ($\pm$7.1e+1)\\
\occlusion&.487 ($\pm$.003)&.298 ($\pm$.006)&6.06e+4 ($\pm$2.9e+2)&.090 ($\pm$.001)&.084 ($\pm$.001)&3.36e+5 ($\pm$2.6e+3)&.760 ($\pm$.004)&.580 ($\pm$.006)&1.40e+4 ($\pm$3.6e+1)\\
\salmean&\textbf{.600 ($\boldsymbol \pm$.002)}&.339 ($\pm$.007)&1.08e+4 ($\pm$5.6e+1)&.114 ($\pm$.005)&.091 ($\pm$.001)&4.28e+3 ($\pm$2.3e+2)&\underline{\textbf{.816 ($\pm$.003)}}&.593 ($\pm$.008)&4.16e+3 ($\pm$1.9e+1)\\
\salnorm&\textbf{.600 ($\boldsymbol \pm$.002)}&.339 ($\pm$.007)&1.06e+4 ($\pm$5.6e+1)&.115 ($\pm$.005)&.090 ($\pm$.001)&4.29e+3 ($\pm$9.9e+1)&.815 ($\pm$.003)&.596 ($\pm$.009)&4.16e+3 ($\pm$1.2e+1)\\
\inputxmean&.435 ($\pm$.001)&.294 ($\pm$.014)&1.07e+4 ($\pm$2.3e+1)&.121 ($\pm$.003)&.086 ($\pm$.002)&4.27e+3 ($\pm$1.8e+2)&.736 ($\pm$.002)&.572 ($\pm$.011)&4.16e+3 ($\pm$1.2e+1)\\
\inputxnorm&.580 ($\pm$.001)&.280 ($\pm$.003)&1.06e+4 ($\pm$6.5e+1)&.113 ($\pm$.004)&.093 ($\pm$.002)&4.09e+3 ($\pm$1.8e+2)&.774 ($\pm$.003)&.501 ($\pm$.006)&4.12e+3 ($\pm$2.7e+1)\\
\guidedmean&\textcolor{bad_res}{.269 ($\pm$.001)}&.299 ($\pm$.017)&1.08e+4 ($\pm$1.7e+2)&\textcolor{bad_res}{.076 ($\pm$.002)}&.086 ($\pm$.002)&4.27e+3 ($\pm$2.2e+2)&\textcolor{bad_res}{.501 ($\pm$.006)}&.573 ($\pm$.013)&4.32e+3 ($\pm$4.0e+2)\\
\guidednorm&\textbf{.600 ($\boldsymbol \pm$.002)}&.339 ($\pm$.007)&1.07e+4 ($\pm$3.4e+1)&.114 ($\pm$.005)&.091 ($\pm$.002)&4.21e+3 ($\pm$2.2e+2)&.815 ($\pm$.003)&.594 ($\pm$.009)&4.14e+3 ($\pm$1.7e+1)\\
\midrule
\multicolumn{10}{c}{\lstm}\\
\shapsamp&.396 ($\pm$.012)&.291 ($\pm$.008)&8.42e+5 ($\pm$1.2e+4)&.086 ($\pm$.001)&.084 ($\pm$.000)&2.30e+8 ($\pm$2.5e+5)&.605 ($\pm$.034)&.588 ($\pm$.020)&1.12e+7 ($\pm$2.1e+6)\\
\lime&.429 ($\pm$.012)&.309 ($\pm$.018)&1.68e+5 ($\pm$2.1e+5)&.089 ($\pm$.001)&.081 ($\pm$.002)&3.00e+8 ($\pm$1.8e+5)&.638 ($\pm$.025)&.588 ($\pm$.021)&5.20e+4 ($\pm$4.1e+3)\\
\occlusion&.358 ($\pm$.003)&.281 ($\pm$.007)&2.46e+5 ($\pm$5.7e+0)&.086 ($\pm$.002)&.083 ($\pm$.002)&1.18e+6 ($\pm$1.1e+3)&.694 ($\pm$.011)&.578 ($\pm$.016)&3.71e+4 ($\pm$2.7e+0)\\
\salmean&.502 ($\pm$.008)&.411 ($\pm$.011)&5.11e+3 ($\pm$6.8e+0)&.108 ($\pm$.001)&.106 ($\pm$.000)&3.04e+3 ($\pm$7.7e+1)&.710 ($\pm$.009)&.546 ($\pm$.000)&1.11e+3 ($\pm$2.8e+0)\\
\salnorm&.502 ($\pm$.008)&.410 ($\pm$.010)&5.12e+3 ($\pm$4.6e+0)&.108 ($\pm$.002)&.106 ($\pm$.002)&3.07e+3 ($\pm$3.9e+1)&.710 ($\pm$.010)&.546 ($\pm$.001)&1.10e+3 ($\pm$1.4e+0)\\
\inputxmean&.364 ($\pm$.004)&.349 ($\pm$.027)&5.12e+3 ($\pm$7.2e+0)&.098 ($\pm$.002)&.096 ($\pm$.002)&3.06e+3 ($\pm$7.0e+1)&\textcolor{bad_res}{.570 ($\pm$.010)}&.601 ($\pm$.017)&1.11e+3 ($\pm$2.2e+0)\\
\inputxnorm&\textbf{.511 ($\pm$.007)}&.389 ($\pm$.004)&5.12e+3 ($\pm$4.2e+0)&\textbf{.110 ($\boldsymbol \pm$.001)}&.107 ($\pm$.000)&3.05e+3 ($\pm$9.9e+1)&.697 ($\pm$.007)&.544 ($\pm$.001)&1.10e+3 ($\pm$1.6e+0)\\
\guidedmean&.333 ($\pm$.009)&.382 ($\pm$.033)&5.11e+3 ($\pm$4.4e+0)&.102 ($\pm$.005)&.098 ($\pm$.003)&3.06e+3 ($\pm$1.0e+2)&\textcolor{bad_res}{.527 ($\pm$.005)}&.570 ($\pm$.031)&1.10e+3 ($\pm$2.2e+0)\\
\guidednorm&.502 ($\pm$.009)&.410 ($\pm$.009)&5.10e+3 ($\pm$2.5e+1)&.109 ($\pm$.001)&.107 ($\pm$.001)&3.08e+3 ($\pm$9.2e+1)&\textbf{.711 ($\boldsymbol \pm$.009)}&.547 ($\pm$.001)&1.10e+3 ($\pm$2.4e+0)\\
\bottomrule
\end{tabular}
\caption{Evaluation of the explainability techniques with Human Agreement (HA) and time for computation. HA is measured with Mean Average Precision (MAP) with the gold human annotations, MAP of a Randomly initialized model (MAP RI). The time is computed with FLOPs. The presented numbers are averaged over five different models and the standard deviation of the scores is presented in brackets. Explainability methods with the best MAP for a particular dataset and model are in bold, while the best MAP across all models for a dataset is underlined as well. Methods that have MAP worse than the randomly generated saliency are in \textcolor{bad_res}{red}.}
\label{tab:human}
\end{table}
\end{landscape}

\begin{table*}[h!]
\centering
\begin{tabular}{lrrr}
\toprule
\textbf{Explain.}&\textbf{e-SNLI}&\textbf{IMDB}&\textbf{TSE}\\
\midrule
\rand&56.05 ($\pm$0.71)&49.26 ($\pm$1.94)&56.45 ($\pm$2.37)\\
\midrule
\multicolumn{4}{c}{\textbf{\trans}}\\
\shapsamp&56.05 ($\pm$0.71)&\textcolor{bad_res}{65.84 ($\pm$11.8)}&52.99 ($\pm$4.24)\\
\lime&48.14 ($\pm$10.8)&\textcolor{bad_res}{59.04 ($\pm$13.7)}& 42.17 ($\pm$7.89)\\
\occlusion&55.24 ($\pm$3.77)&\textcolor{bad_res}{69.00 ($\pm$6.22)}&52.23 ($\pm$4.29)\\
\salmean&37.98 ($\pm$2.18)&\textcolor{bad_res}{49.32 ($\pm$9.01)}&\textbf{39.20 ($\boldsymbol \pm$3.06)}\\
\salnorm&38.01 ($\pm$2.19)&\textbf{49.05 ($\boldsymbol \pm$9.16)}&39.29 ($\pm$3.14)\\
\inputxmean&\textcolor{bad_res}{56.98 ($\pm$1.89)}&\textcolor{bad_res}{64.47 ($\pm$8.70)}&55.52 ($\pm$2.59)\\
\inputxnorm&\textbf{37.05 ($\boldsymbol \pm$2.29)}&\textcolor{bad_res}{50.22 ($\pm$8.85)}&37.04 ($\pm$2.69)\\
\guidedmean&53.43 ($\pm$1.00)&\textcolor{bad_res}{67.68 ($\pm$6.94)}&\textcolor{bad_res}{57.56 ($\pm$2.60)}\\
\guidednorm&38.01 ($\pm$2.19)&\textcolor{bad_res}{49.47 ($\pm$8.89)}&39.26 ($\pm$3.18)\\
\midrule
\multicolumn{4}{c}{\cnn}\\
\shapsamp&51.78 ($\pm$2.24)&\textcolor{bad_res}{59.69 ($\pm$8.37)}&\textcolor{bad_res}{64.72 ($\pm$1.75)}\\
\lime&\textcolor{bad_res}{56.16 ($\pm$1.67)}&\textcolor{bad_res}{59.09 ($\pm$8.48)}&\textcolor{bad_res}{65.78 ($\pm$1.59)}\\
\occlusion&54.32 ($\pm$0.94)&\textcolor{bad_res}{59.86 ($\pm$7.78)}&\textcolor{bad_res}{61.17 ($\pm$1.48)}\\
\salmean&34.26 ($\pm$1.78)&\textcolor{bad_res}{49.61 ($\pm$5.26)}&35.70 ($\pm$2.94)\\
\salnorm&34.16 ($\pm$1.81)&\textbf{49.04 ($\boldsymbol \pm$5.60)}&35.67 ($\pm$2.91)\\
\inputxmean&47.06 ($\pm$3.82)&\textcolor{bad_res}{62.05 ($\pm$7.54)}&\textcolor{bad_res}{64.45 ($\pm$2.99)}\\
\inputxnorm&\underline{\textbf{31.55 ($\boldsymbol \pm$2.83)}}&49.20 ($\pm$5.96)&35.86 ($\pm$3.22)\\
\guidedmean&47.68 ($\pm$2.65)&\textcolor{bad_res}{67.03 ($\pm$4.36)}&44.93 ($\pm$1.57)\\
\guidednorm&34.16 ($\pm$1.81)&\textcolor{bad_res}{49.80 ($\pm$5.99)}&\underline{\textbf{35.60 ($\boldsymbol \pm$2.91)}}\\
\midrule
\multicolumn{4}{c}{\lstm}\\
\shapsamp&51.05 ($\pm$4.47)&44.05 ($\pm$3.06)&53.97 ($\pm$6.00)\\
\lime&51.93 ($\pm$7.73)&\textcolor{bad_res}{44.41 ($\pm$3.04)}&54.95 ($\pm$3.19)\\
\occlusion&54.73 ($\pm$3.12)&45.01 ($\pm$3.84)&48.68 ($\pm$2.28)\\
\salmean&38.29 ($\pm$1.77)&35.98 ($\pm$2.11)&\textbf{37.20 ($\boldsymbol \pm$3.48)}\\
\salnorm&38.26 ($\pm$1.84)&36.22 ($\pm$2.04)&37.23 ($\pm$3.50)\\
\inputxmean&49.52 ($\pm$1.81)&43.57 ($\pm$4.98)&48.71 ($\pm$3.23)\\
\inputxnorm&\textbf{37.95 ($\boldsymbol \pm$2.06)}&36.03 ($\pm$1.97)&36.75 ($\pm$3.35)\\
\guidedmean&44.48 ($\pm$2.12)&46.00 ($\pm$3.20)&43.72 ($\pm$5.69)\\
\guidednorm&38.17 ($\pm$1.80)&\underline{\textbf{35.87 ($\boldsymbol \pm$1.99)}}&37.21 ($\pm$3.48)\\
\bottomrule
\end{tabular}

\caption{Faithfulness-AUC for thresholds $\in$ [0, 10, 20, \dots, 100]. \textit{Lower scores} indicate the ability of the saliency approach to assign higher scores to words more responsible for the final prediction. The presented scores are averaged over the different random initializations and the standard deviation is shown in brackets. Explainability methods with the smallest AUC for a particular dataset and model are in bold, while the smallest AUC across all models for a dataset is underlined as well. Methods that have AUC worse than the randomly generated saliency are in \textcolor{bad_res}{red}.}
\label{tab:faith}
\end{table*}

\begin{landscape}
\begin{table}[p]
\centering
\footnotesize
\begin{tabular}{l@{\hspace{0.2\tabcolsep}}|r@{\hspace{0.7\tabcolsep}}r@{\hspace{0.7\tabcolsep}}r@{\hspace{0.7\tabcolsep}}r@{\hspace{0.5\tabcolsep}}|r@{\hspace{0.7\tabcolsep}}r@{\hspace{0.7\tabcolsep}}r@{\hspace{0.7\tabcolsep}}r@{\hspace{0.5\tabcolsep}}|r@{\hspace{0.7\tabcolsep}}r@{\hspace{0.7\tabcolsep}}r@{\hspace{0.7\tabcolsep}}r}
\toprule
& \multicolumn{4}{c}{\textbf{e-SNLI}}&\multicolumn{4}{c}{\textbf{IMDB}}&\multicolumn{4}{c}{\textbf{TSE}} \\
\textbf{Explain.} & \textbf{MAE} & \textbf{MAX} &\textbf{MAE-up} & \textbf{MAX-up} & \textbf{MAE} & \textbf{MAX} & \textbf{MAE-up} & \textbf{MAX-up} & \textbf{MAE} & \textbf{MAX} & \textbf{MAE-up} & \textbf{MAX-up} \\ \midrule
\rand&.087 ($\pm$.004)&.527 ($\pm$.007)&.276 ($\pm$.005)&.377 ($\pm$.002)&.130 ($\pm$.007)&.286 ($\pm$.014)&.160 ($\pm$.003)&.251 ($\pm$.008)&.092 ($\pm$.009)&.466 ($\pm$.021)&.260 ($\pm$.017)&.428 ($\pm$.064) \\
\midrule
\multicolumn{13}{c}{\textbf{\trans}} \\
\shapsamp&.071 ($\pm$.005)&.456 ($\pm$.037)&.158 ($\pm$.029)&.437 ($\pm$.046)&\textbf{.071 ($\boldsymbol \pm$.008)}&\textbf{.238 ($\boldsymbol \pm$.036)}&\textbf{.120 ($\boldsymbol \pm$.033)}&\textbf{.213 ($\boldsymbol \pm$.035)}&\underline{\textbf{.073 ($\boldsymbol \pm$.012)}}&\textbf{.408 ($\boldsymbol \pm$.043)}&\textbf{.169 ($\boldsymbol \pm$.052)}&\textbf{.415 ($\boldsymbol \pm$.030)} \\
\lime&\underline{\textbf{.068 ($\boldsymbol \pm$.002)}}& \underline{\textbf{.368 ($\boldsymbol \pm$.151)}}&\textbf{.136 ($\boldsymbol \pm$.028)}&\textbf{.395 ($\boldsymbol \pm$.128)}&.077 ($\pm$.008)&.288 ($\pm$.024)&.184 ($\pm$.018)&.260 ($\pm$.021)&.084 ($\pm$.009)&.521 ($\pm$.072)&.232 ($\pm$.013)&.661 ($\pm$.225) \\
\occlusion&.074 ($\pm$.004)&.499 ($\pm$.020)&.224 ($\pm$.006)&.518 ($\pm$.048)&.085 ($\pm$.011)&.306 ($\pm$.015)&.196 ($\pm$.015)&.252 ($\pm$.011)&.085 ($\pm$.011)&.463 ($\pm$.035)&.247 ($\pm$.015)&.482 ($\pm$.091) \\
\salmean&.078 ($\pm$.005)&.544 ($\pm$.014)&.269 ($\pm$.004)&.416 ($\pm$.043)&.083 ($\pm$.009)&.303 ($\pm$.008)&.197 ($\pm$.017)&.269 ($\pm$.023)&.085 ($\pm$.012)&.474 ($\pm$.021)&.248 ($\pm$.017)&.467 ($\pm$.091) \\ 
\salnorm&.078 ($\pm$.005)&.565 ($\pm$.051)&.259 ($\pm$.007)&.571 ($\pm$.095)&.083 ($\pm$.009)&.306 ($\pm$.017)&.195 ($\pm$.021)&.245($\pm$.004)&.085 ($\pm$.012)&.465 ($\pm$.021)&.255 ($\pm$.012)&.479 ($\pm$.074) \\ 
\inputxmean&.079 ($\pm$.005)&.502 ($\pm$.015)&.242 ($\pm$.006)&.518 ($\pm$.031)&.084 ($\pm$.011)&.310 ($\pm$.011)&.198 ($\pm$.013)&.246 ($\pm$.008)&.085 ($\pm$.011)&.463 ($\pm$.015)&.237 ($\pm$.010)&.480 ($\pm$.071) \\
\inputxnorm&.078 ($\pm$.005)&.568 ($\pm$.057)&.258 ($\pm$.007)&.581 ($\pm$.096)&.083 ($\pm$.011)&.301 ($\pm$.014)&.193 ($\pm$.023)&.249 ($\pm$.016)&.086 ($\pm$.013)&.469 ($\pm$.022)&.252 ($\pm$.016)&.480 ($\pm$.087) \\
\guidedmean&.080 ($\pm$.005)&.505 ($\pm$.016)&.242 ($\pm$.008)&.519 ($\pm$.037)&.084 ($\pm$.011)&.308 ($\pm$.009)&.196 ($\pm$.014)&.245 ($\pm$.014)&.085 ($\pm$.011)&.456 ($\pm$.014)&.237 ($\pm$.013)&.494 ($\pm$.069) \\
\guidednorm&.078 ($\pm$.005)&.565 ($\pm$.051)&.258 ($\pm$.007)&.573 ($\pm$.095)&.080 ($\pm$.012)&.306 ($\pm$.009)&.192 ($\pm$.018)&.244 ($\pm$.008)&.086 ($\pm$.012)&.503 ($\pm$.053)&.261 ($\pm$.017)&.450 ($\pm$.081) \\
\midrule
\multicolumn{13}{c}{\textbf{\cnn}} \\
\shapsamp&\textbf{.103 ($\boldsymbol \pm$.001)}&.439 ($\pm$.020)&\textbf{.133 ($\boldsymbol \pm$.003)}&.643 ($\pm$.032)&.077 ($\pm$.018)&.210 ($\pm$.041)&.085 ($\pm$.023)&.196 ($\pm$.026)&.093 ($\pm$.002)&\underline{\textbf{.372 ($\boldsymbol \pm$.011)}}&.148 ($\pm$.004)&.479 ($\pm$.030) \\
\lime&.125 ($\pm$.003)&.498 ($\pm$.018)&.190 ($\pm$.006)&.494 ($\pm$.028)&.128 ($\pm$.006)&.289 ($\pm$.019)&.156 ($\pm$.003)&.260 ($\pm$.011)&.103 ($\pm$.001)&.469 ($\pm$.027)&.202 ($\pm$.014)&.633 ($\pm$.090) \\
\occlusion&.119 ($\pm$.004)&.492 ($\pm$.018)&.176 ($\pm$.007)&.507 ($\pm$.037)&.130 ($\pm$.007)&.289 ($\pm$.018)&.160 ($\pm$.006)&.254 ($\pm$.005)&.114 ($\pm$.002)&.463 ($\pm$.018)&.250 ($\pm$.007)&.418 ($\pm$.035) \\
\salmean&.137 ($\pm$.002)&.496 ($\pm$.011)&.220 ($\pm$.006)&.399 ($\pm$.010)&.129 ($\pm$.007)&.288 ($\pm$.021)&.159 ($\pm$.003)&.253 ($\pm$.013)&.115 ($\pm$.002)&.467 ($\pm$.014)&.245 ($\pm$.007)&.425 ($\pm$.028) \\
\salnorm&.140 ($\pm$.003)&.492 ($\pm$.009)&.225 ($\pm$.005)&.354 ($\boldsymbol \pm$.009)&.130 ($\pm$.006)&.286 ($\pm$.019)&.161 ($\pm$.004)&.250 ($\pm$.005)&.114 ($\pm$.002)&.475 ($\pm$.016)&.248 ($\pm$.006)&.405 ($\pm$.031) \\
\inputxmean&.110 ($\pm$.001)&\textbf{.436 ($\boldsymbol \pm$.014)}&.153 ($\pm$.007)&.460 ($\pm$.009)&\textbf{.071 ($\boldsymbol \pm$.004)}&\underline{\textbf{.191 ($\boldsymbol \pm$.010)}}&\underline{\textbf{.071 ($\boldsymbol \pm$.005)}}&\underline{\textbf{.190 ($\boldsymbol \pm$.010)}}&\textbf{.090 ($\boldsymbol \pm$.002)}&.379 ($\pm$.012)&\underline{\textbf{.135 ($\boldsymbol \pm$.004)}}&.477 ($\boldsymbol \pm$.025) \\ 
\inputxnorm&.140 ($\pm$.003)&.492 ($\pm$.009)&.225 ($\pm$.005)&.355 ($\pm$.007)&.130 ($\pm$.007)&.285 ($\pm$.019)&.160 ($\pm$.004)&.251 ($\pm$.011)&.114 ($\pm$.002)&.475 ($\pm$.014)&.248 ($\pm$.006)&.416 ($\pm$.033) \\
\guidedmean&.140 ($\pm$.003)&.485 ($\pm$.011)&.225 ($\pm$.005)&.367 ($\pm$.023)&.129 ($\pm$.006)&.286 ($\pm$.019)&.159 ($\pm$.003)&.253 ($\pm$.011)&.114 ($\pm$.002)&.462 ($\pm$.013)&.234 ($\pm$.011)&.441 ($\pm$.036) \\
\guidednorm&.140 ($\pm$.003)&.492 ($\pm$.009)&.225 ($\pm$.005)&\underline{\textbf{.353 ($\boldsymbol \pm$.008)}}&.130 ($\pm$.007)&.289 ($\pm$.018)&.159 ($\pm$.004)&.252 ($\pm$.011)&.114 ($\pm$.002)&.473 ($\pm$.015)&.249 ($\pm$.006)&\textbf{.404 ($\boldsymbol \pm$.029)} \\
\midrule
\multicolumn{13}{c}{\textbf{\lstm}} \\
\shapsamp&\textbf{.118 ($\boldsymbol \pm$.003)}&.622 ($\boldsymbol \pm$.035)&\underline{\textbf{.131 ($\boldsymbol \pm$.005)}}&.648 ($\pm$.054)&\underline{\textbf{.060 ($\boldsymbol \pm$.018)}}&\textbf{.279 ($\boldsymbol \pm$.065)}&\textbf{.160 ($\boldsymbol \pm$.014)}&.277 ($\pm$.038)&\textbf{.087 ($\boldsymbol \pm$.007)}&\textbf{.433 ($\boldsymbol \pm$.053)}&\textbf{.147 ($\boldsymbol \pm$.015)}&\underline{\textbf{.393 ($\boldsymbol \pm$.029)}}\\
\lime&.127 ($\pm$.004)&.512 ($\pm$.052)&.145 ($\pm$.009)&.490 ($\pm$.040)&.069 ($\pm$.018)&.300 ($\pm$.051)&.209 ($\pm$.024)&.267 ($\pm$.031)&.090 ($\pm$.007)&.667 ($\pm$.150)&.218 ($\pm$.010)&.864 ($\pm$.362) \\
\occlusion&.147 ($\pm$.003)&.579 ($\pm$.065)&.172 ($\pm$.007)&.593 ($\pm$.083)&.069 ($\pm$.017)&.304 ($\pm$.055)&.216 ($\pm$.014)&.324 ($\pm$.032)&.099 ($\pm$.006)&.509 ($\pm$.015)&.259 ($\pm$.012)&.723 ($\pm$.063) \\
\salmean&.163 ($\pm$.002)&.450 ($\pm$.008)&.195 ($\pm$.008)&.398 ($\pm$.031)&.069 ($\pm$.018)&.301 ($\pm$.051)&.208 ($\pm$.026)&\textbf{.259 ($\boldsymbol \pm$.022)}&.101 ($\pm$.007)&.518 ($\pm$.013)&.271 ($\pm$.008)&.469 ($\pm$.071) \\
\salnorm&.163 ($\pm$.002)&.448 ($\pm$.011)&.195 ($\pm$.008)&.399 ($\pm$.034)&.070 ($\pm$.018)&.299 ($\pm$.051)&.206 ($\pm$.024)&.263 ($\pm$.027)&.101 ($\pm$.007)&.523 ($\pm$.011)&.273 ($\pm$.008)&.441 ($\pm$.051) \\
\inputxmean&.161 ($\pm$.002)&.454 ($\pm$.018)&.193 ($\pm$.007)&.502 ($\pm$.033)&.066 ($\pm$.018)&.295 ($\pm$.059)&.201 ($\pm$.033)&\textbf{.262 ($\boldsymbol \pm$.014)}&.098 ($\pm$.007)&.527 ($\pm$.005)&.268 ($\pm$.008)&.425 ($\pm$.035) \\
\inputxnorm&.163 ($\pm$.002)&\textbf{.445 ($\boldsymbol \pm$.011)}&.195 ($\pm$.007)&\textbf{.394 ($\boldsymbol \pm$.029)}&.068 ($\pm$.018)&.303 ($\pm$.050)&.201 ($\pm$.031)&.277 ($\pm$.024)&.101 ($\pm$.007)&.523 ($\pm$.008)&.273 ($\pm$.007)&.445 ($\pm$.038) \\
\guidedmean&.161 ($\pm$.001)&.453 ($\pm$.014)&.192 ($\pm$.007)&.516 ($\pm$.058)&.068 ($\pm$.019)&.298 ($\pm$.055)&.200 ($\pm$.024)&.287 ($\pm$.045)&.097 ($\pm$.006)&.523 ($\pm$.017)&.260 ($\pm$.016)&.460 ($\pm$.045) \\
\guidednorm&.163 ($\pm$.002)&.446 ($\pm$.010)&.195 ($\pm$.007)&.396 ($\pm$.042)&.069 ($\pm$.017)&.300 ($\pm$.050)&.204 ($\pm$.024)&.279 ($\pm$.025)&.101 ($\pm$.007)&.525 ($\pm$.010)&.273 ($\pm$.007)&.474 ($\pm$.051) \\
\bottomrule
\end{tabular}
\caption{Confidence Indication experiments are measured with the Mean Absolute Error (MAE) of the generated saliency scores when used to predict the confidence of the class predicted by the model and the Maximum Error (MAX). We present the result with and without up-sampling(MAE-up, MAX-up) of the model confidence.
The presented measures are an average over the set of models trained from from different random seeds. The standard deviation of the scores is presented in brackets.  AVG Conf. is the average confidence of the model for the predicted class. The best results for a particular dataset and model are in bold and the best results across a dataset are also underlined. Lower results are better.}
~\label{tab:confidence}
\end{table}
\end{landscape}

\begin{table*}[h!]
\centering
\begin{tabular}{lrrr}
\toprule
\textbf{Explain.} & \textbf{e-SNLI} & \textbf{IMDB} & \textbf{TSE} \\
\midrule
\multicolumn{4}{c}{\textbf{\trans}} \\
\rand & -0.004 (2.6e-01)    & -0.035 (1.4e-01)  & 0.003 (6.1e-01) \\
\shapsamp & 0.310 (0.0e+00) & 0.234 (3.6e-12)   & 0.259 (0.0e+00) \\
\lime & \textbf{0.519 (0.0e+00)} & 0.269 (3.0e-31) & 0.110 (2.0e-29) \\
\occlusion & 0.215 (0.0e+00) & 0.341 (2.6e-50) & 0.255 (0.0e+00) \\
\salmean & 0.356 (0.0e+00) & 0.423 (3.9e-79) & \textbf{0.294 (0.0e+00)} \\
\salnorm & 0.297 (0.0e+00) & 0.405 (6.9e-72) & 0.289 (0.0e+00) \\
\inputxmean & \textcolor{bad_res}{-0.102 (2.0e-202)} & \textbf{0.426 (2.5e-80)} & \textcolor{bad_res}{-0.010 (1.3e-01)} \\
\inputxnorm & 0.311 (0.0e+00) & 0.397 (3.8e-69) & 0.292 (0.0e+00) \\
\guidedmean & 0.064 (1.0e-79) & \textcolor{bad_res}{-0.083 (4.2e-04)} & \textcolor{bad_res}{-0.005 (4.9e-01)} \\
\guidednorm & 0.297 (0.0e+00) & 0.409 (1.2e-73) & 0.293 (0.0e+00) \\
\midrule
\multicolumn{4}{c}{\textbf{\cnn}} \\
\rand & -0.003 (4.0e-01) & 0.426 (2.6e-106) & -0.002 (7.4e-01) \\
\shapsamp & 0.789 (0.0e+00) & 0.537 (1.4e-179) & 0.704 (0.0e+00) \\
\lime & 0.790 (0.0e+00) & 0.584 (1.9e-219) & \underline{\textbf{0.730 (0.0e+00)}} \\
\occlusion & 0.730 (0.0e+00) & 0.528 (2.4e-172) & 0.372 (0.0e+00) \\
\salmean & 0.701 (0.0e+00) & 0.460 (4.5e-126) & 0.320 (0.0e+00) \\
\salnorm & \underline{\textbf{0.819 (0.0e+00)}} & 0.583 (4.0e-218) & 0.499 (0.0e+00) \\
\inputxmean & 0.136 (0.0e+00) & \textcolor{bad_res}{0.331 (1.2e-62)} & 0.002 (7.5e-01) \\
\inputxnorm & 0.816 (0.0e+00) & \underline{\textbf{0.585 (8.6e-221)}} & 0.495 (0.0e+00) \\
\guidedmean & 0.160 (0.0e+00) &\textcolor{bad_res}{ 0.373 (5.5e-80)} & 0.173 (6.3e-121) \\
\guidednorm & \underline{\textbf{0.819 (0.0e+00)}} & 0.578 (2.4e-214) & 0.498 (0.0e+00) \\
\midrule
\multicolumn{4}{c}{\textbf{\lstm}} \\ 
\rand & 0.004 (1.8e-01) & 0.002 (9.2e-01) & 0.010 (1.8e-01) \\
\shapsamp & 0.657 (0.0e+00) & 0.382 (1.7e-63) & 0.502 (0.0e-00) \\
\lime & \textbf{0.700 (0.0e+00)} & 0.178 (3.3e-14) & 0.540 (0.0e-00) \\
\occlusion & 0.697 (0.0e+00) & \textbf{0.498 (1.7e-113)} & 0.454 (0.0e-00) \\
\salmean & 0.645 (0.0e+00) & 0.098 (3.1e-05) & \textbf{0.667 (0.0e-00)} \\
\salnorm & 0.662 (0.0e+00) & 0.132 (1.8e-08) & 0.596 (0.0e-00) \\
\inputxmean & 0.026 (1.9e-14) & \textcolor{bad_res}{-0.032 (1.7e-01)} & 0.385 (0.0e-00) \\
\inputxnorm & 0.664 (0.0e+00) & 0.133 (1.5e-08) & 0.604 (0.0e-00) \\
\guidedmean & 0.144 (0.0e+00) & 0.122 (2.0e-07) & 0.295 (0.0e-00) \\
\guidednorm & 0.663 (0.0e+00) & 0.139 (3.1e-09) & 0.598 (0.0e-00) \\
\bottomrule
\end{tabular}
\caption{Rationale Consistency Spearman's $\rho$ correlation. The estimated p-value for the correlation is provided in the brackets. The best results for a particular dataset and model are in bold and the best results across a dataset are also underlined. Correlation lower that the one of the randomly sampled saliency scores are colored in \textcolor{bad_res}{red}.}
\label{tab:consistency:rat}
\end{table*}

\begin{table*}[h!]
\centering
\begin{tabular}{llll}
\toprule
\textbf{Explain.} & \textbf{e-SNLI} & \textbf{IMDB} & \textbf{TSE} \\
\midrule
\multicolumn{4}{c}{\textbf{\trans}}  \\
\rand & 0.047 (2.7e-04) & 0.127 (6.6e-07)/ & 0.121 (2.5e-01) \\
\shapsamp & 0.285 (1.8e-02) & \textcolor{bad_res}{0.078 (5.8e-04)} & 0.308 (3.4e-36) \\
\lime & 0.372 (3.1e-90) & \textbf{0.236 (4.6e-07)} & \underline{\textbf{0.413 (3.4e-120)}} \\
\occlusion & 0.215 (9.6e-02) & \textcolor{bad_res}{0.003 (2.0e-04)} & 0.235 (7.3e-05) \\
\salmean & 0.378 (4.3e-57) & \textcolor{bad_res}{0.023 (4.3e-02)} & 0.253 (1.4e-20) \\
\salnorm & 0.027 (3.0e-05) & \textcolor{bad_res}{-0.043 (5.6e-02)} & 0.260 (6.8e-21) \\
\inputxmean & 0.319 (3.0e-03) & \textcolor{bad_res}{0.008 (1.2e-01)} & 0.193 (7.5e-05) \\
\inputxnorm & 0.399 (1.9e-78) & \textcolor{bad_res}{0.028 (2.3e-03)} & 0.247 (4.9e-17) \\
\guidedmean & 0.400 (6.7e-31) & \textcolor{bad_res}{0.017 (1.9e-01)} & 0.228 (5.2e-09) \\
\guidednorm & \underline{\textbf{0.404 (1.4e-84)}} & \textcolor{bad_res}{0.019 (4.3e-04)} & 0.255 (3.1e-20) \\
\midrule
\multicolumn{4}{c}{\textbf{\cnn}} \\
\rand & 0.018 (2.4e-01) & 0.115 (1.8e-04) & 0.008 (2.0e-01) \\
\shapsamp & \textcolor{bad_res}{0.015 (1.8e-01)} & \textcolor{bad_res}{-0.428 (5.3e-153)} & 0.037 (1.4e-01) \\
\lime & \textcolor{bad_res}{0.000 (4.4e-02)} & 0.400 (1.4e-126) & 0.023 (4.0e-01) \\
\occlusion & \textcolor{bad_res}{-0.076 (6.5e-02)} & \textcolor{bad_res}{-0.357 (1.9e-85)} & \textbf{0.041 (1.7e-01)} \\
\salmean & 0.381 (6.9e-91) & 0.431 (1.1e-146) & \textcolor{bad_res}{-0.100 (3.9e-06)} \\
\salnorm & 0.391 (1.7e-98) & 0.427 (3.5e-135) & \textcolor{bad_res}{-0.100 (3.7e-06)} \\
\inputxmean & 0.171 (5.1e-04) & 0.319 (1.4e-69) & 0.024 (3.5e-01) \\
\inputxnorm & \textbf{0.399 (1.0e-93)} & 0.428 (1.4e-132) & \textcolor{bad_res}{-0.076 (1.2e-03)} \\
\guidedmean & 0.091 (7.9e-02) & 0.375 (5.7e-109) & \textcolor{bad_res}{-0.032 (1.1e-01)} \\
\guidednorm & \textbf{0.391 (1.7e-98)} & \underline{\textbf{0.432 (3.5e-140)}} & \textcolor{bad_res}{-0.102 (1.7e-06)} \\
\midrule
\multicolumn{4}{c}{\textbf{\lstm}} \\
\rand & 0.018 (3.9e-01) & 0.037 (1.8e-01) & 0.016 (9.2e-03) \\
\shapsamp & 0.398 (3.5e-81) & 0.230 (8.9e-03) & 0.205 (2.1e-16) \\
\lime & \underline{\textbf{0.415 (1.2e-80)}} & 0.079 (8.6e-04) & 0.207 (4.3e-16) \\
\occlusion & 0.363 (1.1e-37) & \textbf{0.429 (7.5e-137)} & \textbf{0.237 (2.9e-29)} \\
\salmean & 0.158 (1.7e-17) & \textcolor{bad_res}{-0.177 (1.6e-10)} & 0.065 (5.8e-03) \\
\salnorm & 0.160 (7.5e-19) & \textcolor{bad_res}{-0.168 (2.0e-15)} & 0.096 (8.2e-03) \\
\inputxmean & 0.142 (3.3e-06) & \textcolor{bad_res}{-0.152 (1.2e-14)} & 0.106 (2.8e-02) \\
\inputxnorm & 0.183 (7.0e-24) & \textcolor{bad_res}{-0.175 (4.7e-17)} & 0.089 (8.4e-03) \\
\guidedmean & 0.163 (1.9e-12) & \textcolor{bad_res}{-0.060 (4.7e-02)} & 0.077 (1.2e-02) \\
\guidednorm & 0.169 (1.8e-12) & \textcolor{bad_res}{-0.214 (5.8e-16)} & 0.115 (4.3e-02) \\
\bottomrule
\end{tabular}
\caption{Dataset Consistency results with Spearman $\rho$. The estimated p-value for the correlation is provided in the brackets. The best results for a particular dataset and model are in bold and the best results across a dataset are also underlined. Correlation lower that the one of the randomly samples saliency scores are colored in \textcolor{bad_res}{red}.}
\label{tab:consistency:data}
\end{table*}

\end{document}